\documentclass[10pt,twocolumn,letterpaper]{article}

\usepackage{cvm}
\usepackage{times}
\usepackage{epsfig}
\usepackage{graphicx}
\usepackage{amsmath}
\usepackage{amssymb}
\usepackage{booktabs}
\usepackage{multirow}
\usepackage{adjustbox}

\usepackage[pagebackref=true,breaklinks=true,letterpaper=true,colorlinks,bookmarks=false]{hyperref}

\cvmfinalcopy

\def\cvmPaperID{****} 

\ifcvmfinal\fi
\begin{document}
\newcommand{\sfm}{\mathrm{SFM}}
\title{Point cloud completion via structured feature maps using a feedback network}

\author{Zejia Su\\
Shenzhen University\\
Shenzhen, China\\
{\tt\small zejiasu.36@gmail.com}
\and
Haibin Huang\\
Kuaishou Technology\\
Beijing, China\\
{\tt\small jackiehuanghaibin@gmail.com}
\and
Chongyang Ma\\
Kuaishou Technology\\
Beijing, China\\
{\tt\small chongyangm@gmail.com}
\and
Hui Huang\\
Shenzhen University\\
Shenzhen, China\\
{\tt\small hhzhiyan@gmail.com}
\and
Ruizhen Hu\\
Shenzhen University\\
Shenzhen, China\\
{\tt\small ruizhen.hu@gmail.com}
}

\maketitle

\begin{abstract}
    In this paper, we tackle the challenging problem of point cloud completion from the perspective of feature learning. Our key observation is that to recover the underlying structures as well as surface details, given partial input, a fundamental component is a good feature representation that can capture both global structure and local geometric details. We accordingly first  propose FSNet, a feature structuring module that can adaptively aggregate point-wise features into a 2D structured feature map by learning multiple latent patterns from local regions. We then integrate FSNet into a coarse-to-fine pipeline for point cloud completion. Specifically, a 2D convolutional neural network is adopted to decode feature maps from FSNet into a coarse and complete point cloud. Next, a point cloud upsampling network is used to generate a dense point cloud from the partial input and the coarse intermediate output. To efficiently exploit  local structures and enhance point distribution uniformity, we propose IFNet, a point upsampling module with a self-correction mechanism that can progressively refine details of the generated dense point cloud. We have conducted qualitative and quantitative experiments on ShapeNet, MVP, and KITTI datasets, which demonstrate that our method outperforms state-of-the-art point cloud completion approaches.
\end{abstract}

\section{Introduction}

In this paper, we study the problem of point cloud completion, i.e.\ recovering a full point cloud given a partial observation.  It is an important component of many real-world applications, such as 3D data scanning~\cite{armeni20163d}, acquisition~\cite{tarini20053d}, robot navigation~\cite{kruse2013human} and so on. 

\begin{figure*}
\centering
\begin{tabular}{ccccc}
Input & PMP-Net & SnowflakeNet & Ours & GT \\
\includegraphics[width=0.178\textwidth]{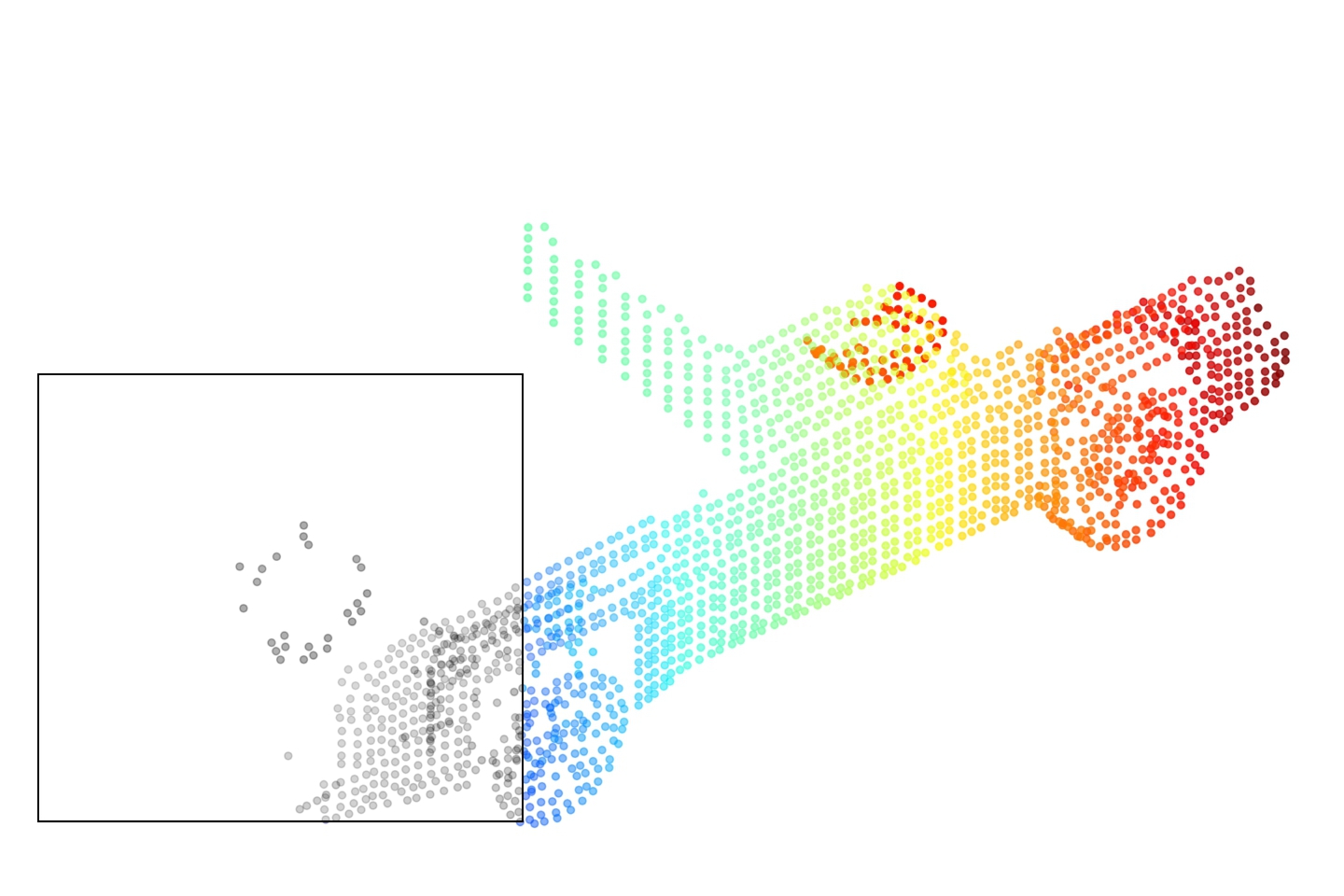} & 
\includegraphics[width=0.178\textwidth]{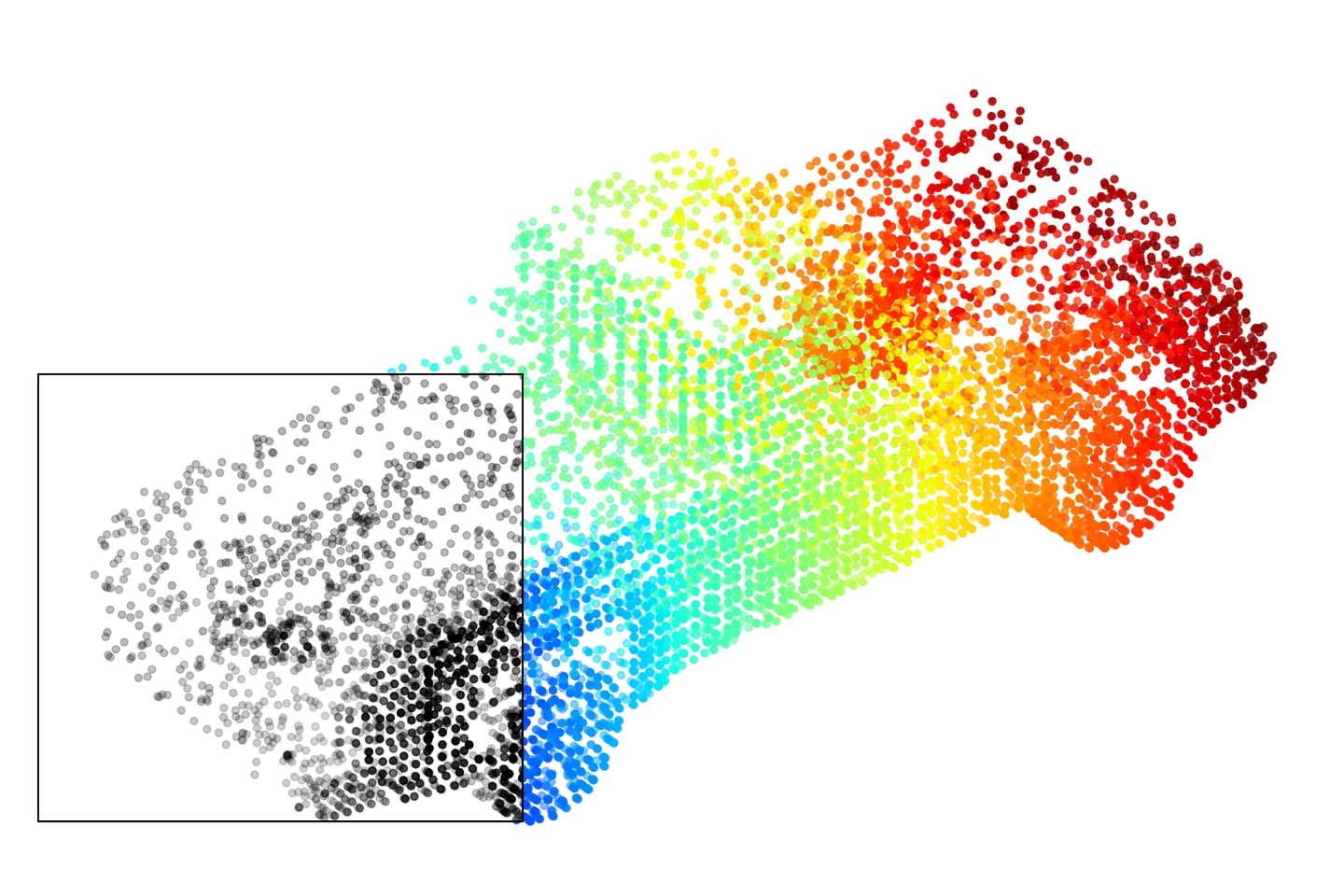} & 
\includegraphics[width=0.178\textwidth]{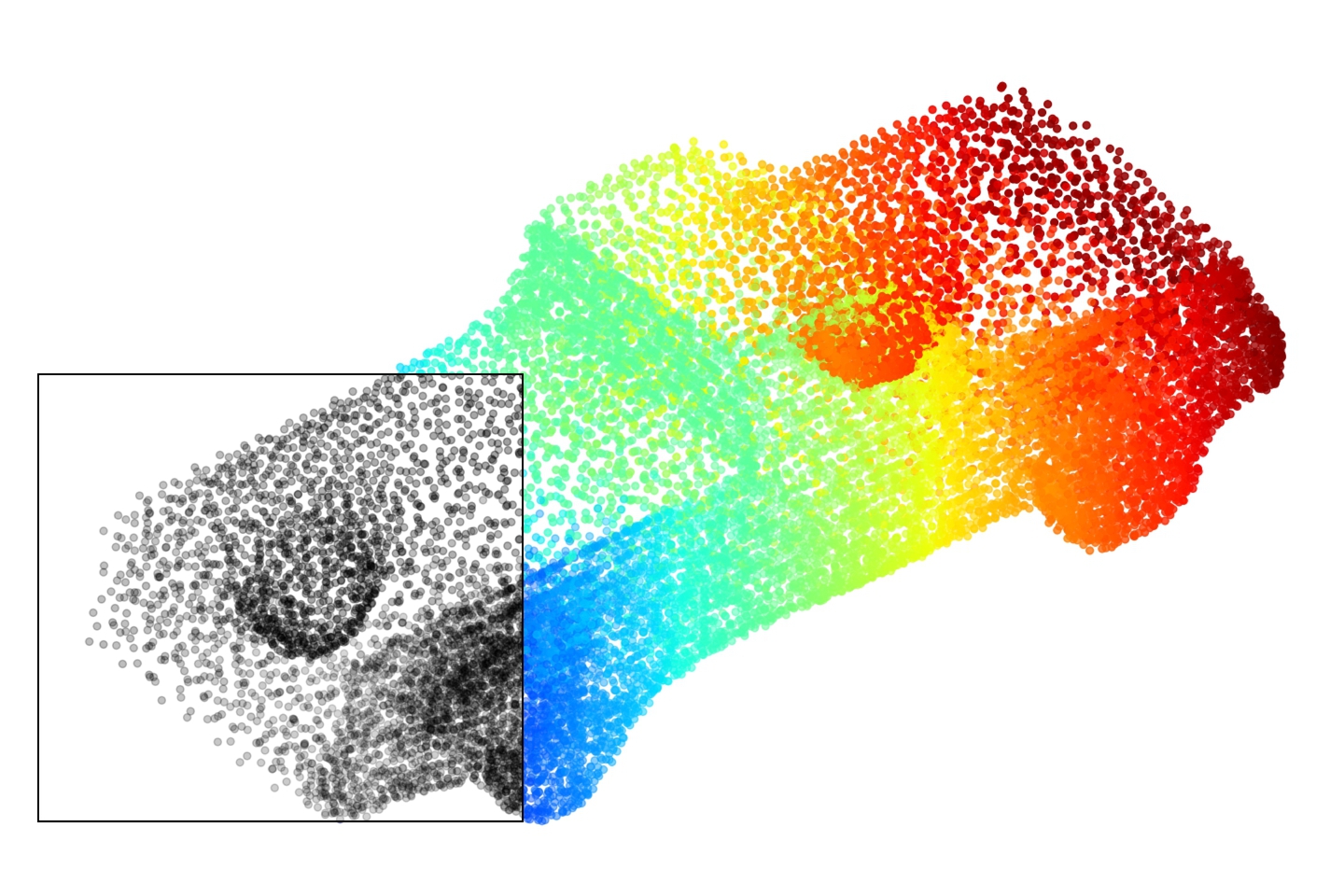} & 
\includegraphics[width=0.178\textwidth]{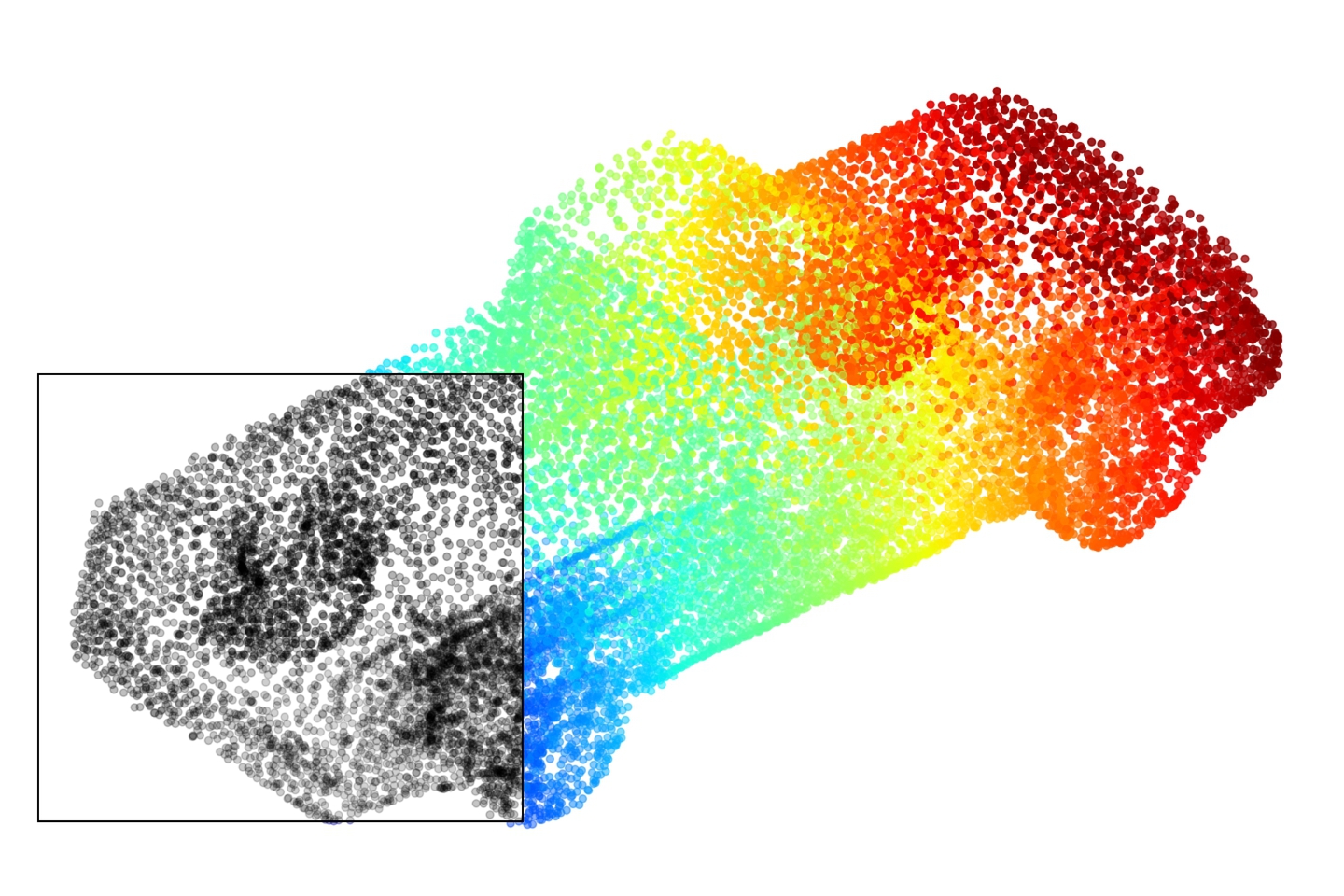} & 
\includegraphics[width=0.178\textwidth]{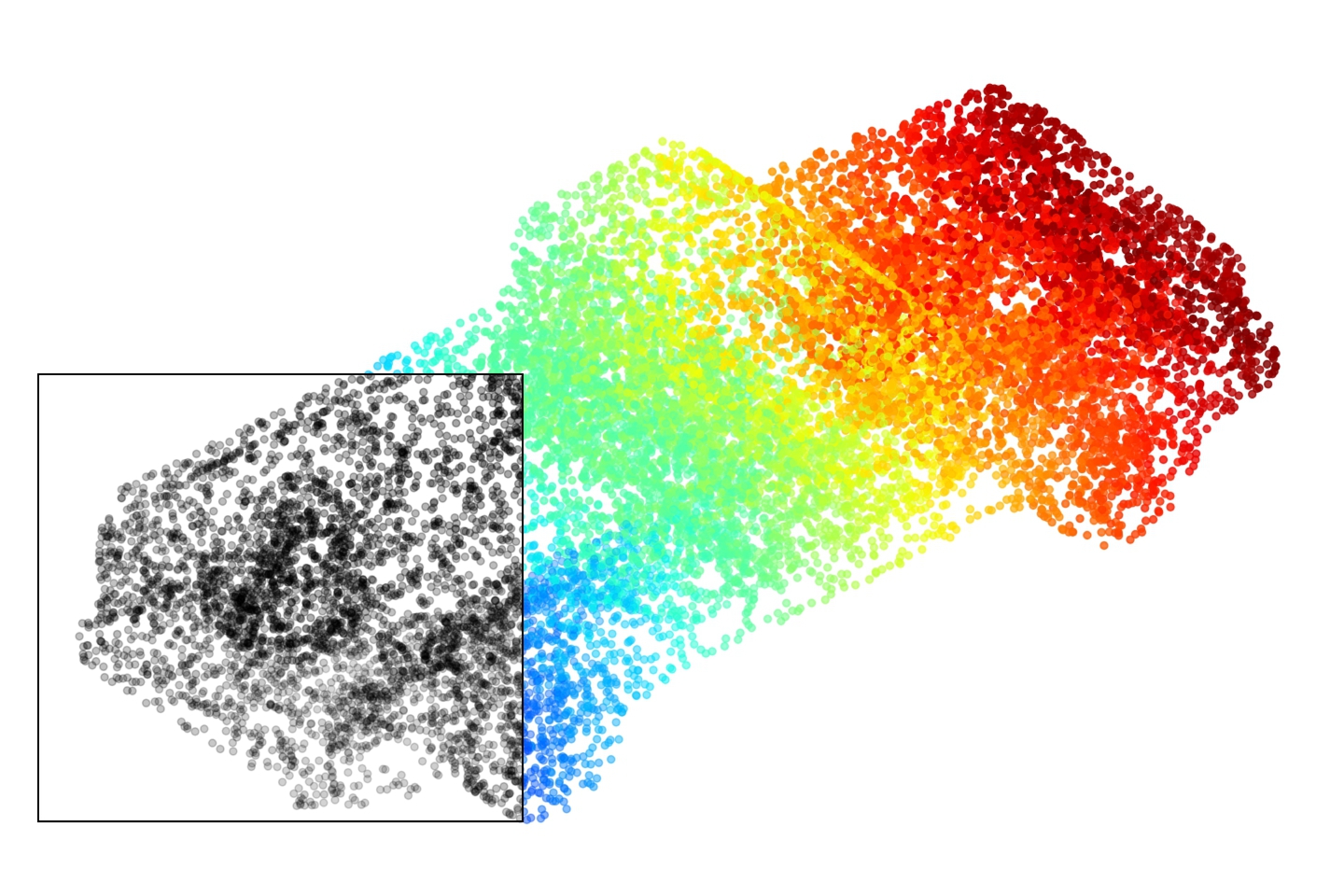}\\
\includegraphics[width=0.178\textwidth]{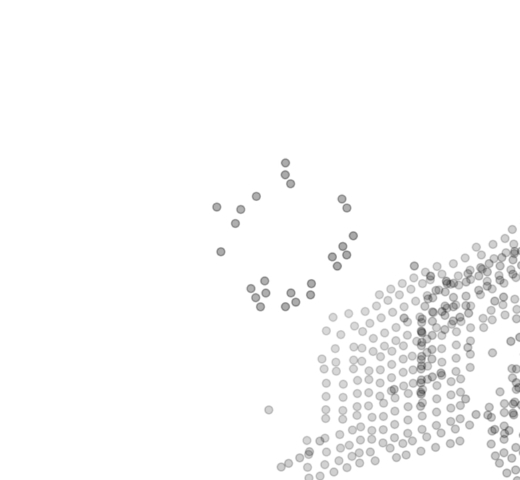} & 
\includegraphics[width=0.178\textwidth]{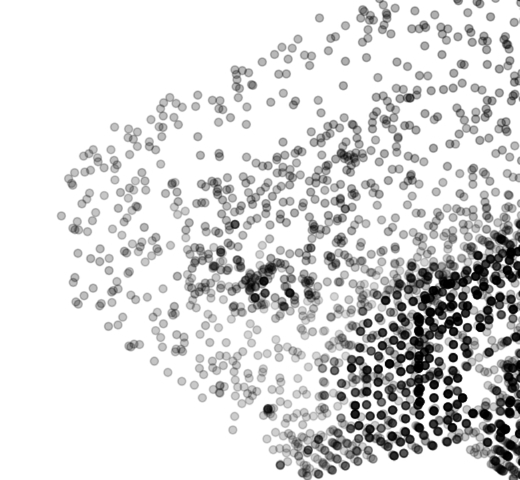} & 
\includegraphics[width=0.178\textwidth]{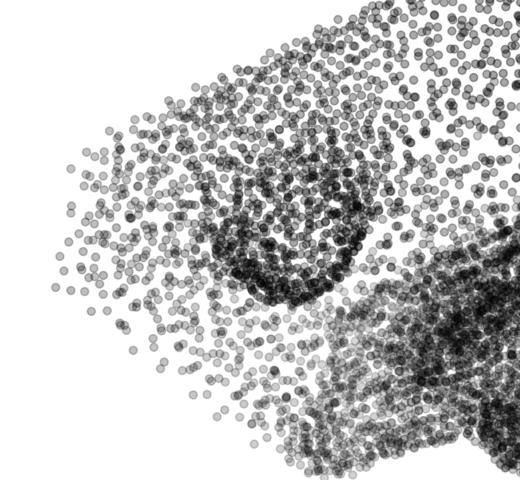} & 
\includegraphics[width=0.178\textwidth]{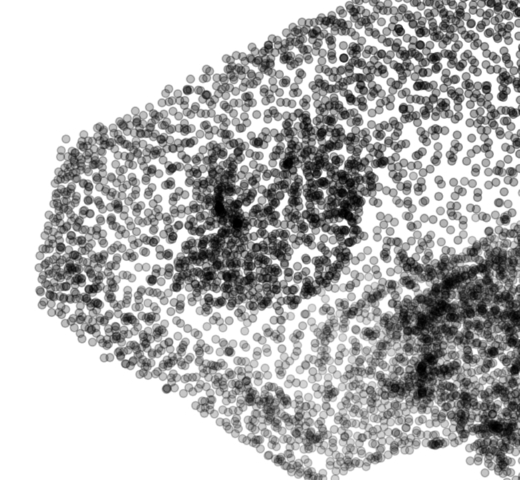} & 
\includegraphics[width=0.178\textwidth]{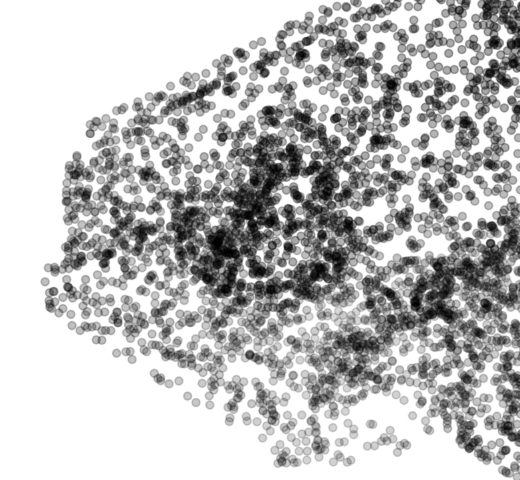}
\end{tabular}
\caption{Our approach can generate complete point clouds from partial input with finer details and more uniform surface points compared to state-of-the-art methods.}
\label{fig9}
\label{fig:teaser}
\end{figure*}

Point cloud completion is a challenging task as it needs to recover both  missing topological structures and geometric details from incomplete input.  Traditional methods~\cite{berger2014state, mitra2013symmetry} used hand-crafted features like surface smoothness or symmetry priors. Such empirical human-designed features suffer from performance degradation under changing illumination and in the presence of severe occlusion.
Recently, deep neural network (DNN) based methods have been introduced into this task and have achieved promising improvements. Early works~\cite{dai2017shape,stutz2018learning} voxelized the 3D point cloud and applied a 3D convolutional neural network (CNN) to the volumetric data to complete the shapes. Due to the large computational cost of 3D CNNs, such methods generally output shapes with limited details. Pioneered by PointNet~\cite{qi2017pointnet} and PointNet++~\cite{qi2017pointnet++}, other methods~\cite{yuan2018pcn,tchapmi2019topnet} directly complete  3D point clouds. Compared to using voxels, learning on 3D points is more scalable and efficient, but also leads to the problem of feature learning due to the inherent irregularity and sparseness of point clouds. 

A common practice in previous methods is to apply MLPs to point clouds in an encoder-decoder manner, where per-point features are aggregated to a global feature vector via a max-pooling operation and then decoded into a complete point cloud. Due to the use of pooling operations, these methods suffer from unavoidable information loss which results in unsatisfactory shape structures and blurred details in the missing regions. 
More recent methods have been proposed to improve feature learning, including use of shape priors~\cite{pan2021variational}, a skip-attention mechanism~\cite{wen2020point}, and so on~\cite{zhang2020detail, xie2020grnet}.
However, most of these methods adopt feature vectors as the global representation and suffer from non-uniform distributions of the dense prediction results, due to the irregular and discrete nature of the point cloud, especially when the input shapes have large missing parts. As shown in Figure~\ref{fig:teaser}, the distributions of the point clouds predicted by previous methods are uneven \eg in regions between the known and predicted parts.

To tackle these challenges, we propose a new point cloud completion method with two novel modules: FSNet for structured feature learning, which captures well both global structure and local geometric details of the input point cloud, and IFNet, for progressive detail generation and uniformity enhancement.
Given a  partial input, point-wise features are first extracted by an encoder and FSNet is then applied to form a 2D structured feature map.  Instead of a max-pooling operation, FSNet relies on multi-head attention~\cite{vaswani2017attention} to aggregate the features, where a learnable set and the input features are used as queries and key-value pairs, respectively, and the output matrices constitute the structured feature map.
By exploiting latent patterns among local regions, FSNet can learn a global representation with rich semantic information which is distinguishable at the instance level. The 2D structured feature map then allows us to utilize a 2D convolutional neural network as a decoder to predict the complete structures and generate a coarse point cloud as the input to the upsampling stage. Motivated by the success of feedback mechanisms in the image super-resolution task~\cite{haris2018deep}, we propose IFNet with a self-correction strategy to upsample the coarse point cloud in a progressive manner. A sparse encoding module is applied to extract sparse features from the coarse point cloud and input, and followed by an upsampling module to generate initial dense features. The final dense point cloud is then iteratively refined through IFNet, where  details are progressively added and  uniformity is gradually improved by feeding back the projection error to the initial dense features.
 
We have conducted experiments on ShapeNet, MVP and KITTI datasets to evaluate our method. The results show that our approach can outperform state-of-the-art methods. 

To summarize, our contributions are as follows:
\begin{itemize}
\item FSNet, a novel feature aggregation network which adaptively organizes the unordered features into a 2D structured feature map, which can retain more information and represent more fine-grained global features than a feature vector,
\item IFNet, an iterative feedback network to upsample and refine the coarse completion via a multi-step self-correction procedure, which can further exploit local details and facilitate consolidation and uniformity of the point cloud, and  
\item an evaluation of different algorithms on both synthetic and real-world datasets, which shows that our approach outperforms state-of-the-art methods in terms of both completion quality and surface uniformity.  
\end{itemize}

\begin{figure*}
\centering
\includegraphics[width=\textwidth]{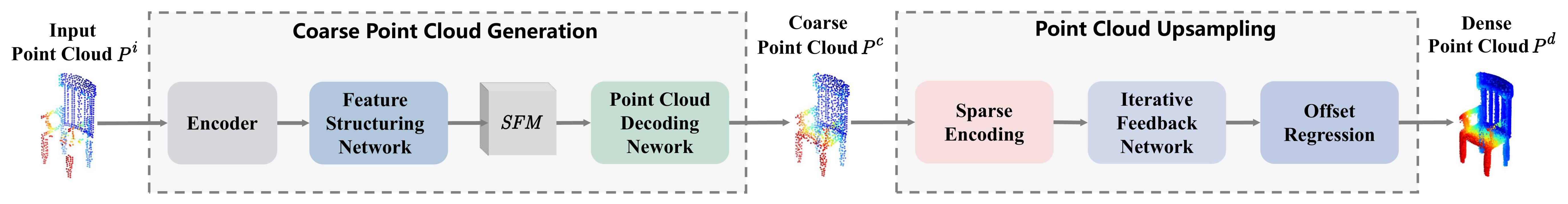}
\caption{Pipeline.}
\label{fig:overall_pipeline}
\end{figure*}

\section{Related Work}
\subsection{3D shape completion}
3D shape completion has drawn increasing attention in recent years. Early geometric methods~\cite{berger2014state, mitra2013symmetry} complete objects by leveraging predefined geometric features, such as surface smoothness or symmetry priors. Since these methods rely on hand-crafted features, they are only valid under particular circumstances. Recently, convolutional neural networks (CNNs) have been widely used for processing regularly arranged data due to their ability to learn features. Some methods~\cite{dai2017shape,stutz2018learning} introduce 3D voxel grids or distance fields as the representation for 3D data, and utilize a 3D CNN to complete objects. However, use of a 3D CNN leads to large computational and memory costs that are cubic in the resolution of the volumetric data, while reducing the resolution limits the processing of fine-grained shapes.

Recent works tend to adopt 3D point clouds as the representations of 3D objects due to their convenience and flexibility. Pioneered by Pointnet~\cite{qi2017pointnet}, several methods use MLPs for point cloud completion under an encoder-decoder framework, where per-point features are aggregated to a global feature vector (GFV) by the max-pooling operation. PCN~\cite{yuan2018pcn} and MSN~\cite{liu2020morphing} complete the point cloud in a coarse-to-fine fashion. TopNet~\cite{tchapmi2019topnet} introduces a hierarchical tree-structure network that takes the geometric structure into consideration. RL-GAN-Net~\cite{sarmad2019rl} and Render4Completion~\cite{hu2019render4completion} focus on  adversarial learning to improve the realism and consistency of the generated shape. However, these methods suffer from  loss of structural details, as they predict the point cloud only from a single global vector.

To better preserve shape structures and complete surface details, SA-Net~\cite{wen2020point} and ASHF-Net~\cite{zong2021ashf} introduce a skip-attention mechanism to further revisit the low-level features from the encoder. SoftpoolNet~\cite{wang2020softpoolnet} proposes a soft pooling module to replace the max-pooling operator, to keep more information by considering multiple features. NSFA~\cite{zhang2020detail} aggregates different features to represent the known and  missing parts separately.
GRNet~\cite{xie2020grnet} and VE-PCN~\cite{wang2021voxel} introduce 3D voxels as the intermediate representation to help the network to infer the the complete shape.
PF-Net~\cite{huang2020pf}, DeCo~\cite{alliegro2021denoise}, and PoinTr~\cite{yu2021pointr} only generate the missing part of the object to preserve the spatial arrangements of the original part. 
VRCNet~\cite{pan2021variational} and ASFM-Net~\cite{xia2021asfm} improve the global features by narrowing the distribution difference between the incomplete and complete point clouds. 
CRN~\cite{wang2020cascaded} and SnowflakeNet~\cite{xiang2021snowflakenet} generate the complete point cloud in a progressive manner.
Other notable work such as PMP-Net~\cite{wen2021pmp} formulates  completion as a point cloud deformation process, where point-wise paths are predicted to move each point of the incomplete input to complete the point cloud. SpareNet~\cite{xie2021style} proposes a style-based point generator with adversarial rendering for point cloud completion.
Cycle4Completion~\cite{wen2021cycle4completion} improves completion quality by establishing a geometric correspondence between complete shapes and incomplete ones.

\subsection{Point cloud upsampling}
Point cloud upsampling aims to generate a uniform dense point cloud to represent the underlying surface of the object within local patches. PU-Net~\cite{yu2018pu} uses PointNet++~\cite{qi2017pointnet++} to extract multi-scale features and expands the features through multi-branch MLPs. With additional edge and surface annotations, EC-Net~\cite{yu2018ec} improves the edge quality of PU-Net by formulating an edge-aware joint loss to learn the geometry of edges. MPU~\cite{yifan2019patch} proposes a progressive network to upsample points in multiple stages, which requires supervision of intermediate outputs. PU-GAN~\cite{li2019pu} presents a generative adversarial network (GAN) to learn the distribution of dense point clouds. PUGeo-Net~\cite{qian2020pugeo} proposes a geometric-centric network by learning local parameterization and normal direction for each point, which needs additional normal annotations. Recently, PU-GCN~\cite{qian2021pu} designed a novel feature expansion network called NodeShuffle, which utilizes graph convolutional networks (GCNs) to encode local information. 

\section{Method}
We now formally introduce our proposed method.
As shown in Figure~\ref{fig:overall_pipeline}, we use a two-stage pipeline which works in a coarse-to-fine manner and consists of a coarse structure completion stage and a point cloud upsampling stage.
More specifically, taking a partial point cloud ${P^i}\in\mathbb{R}^{N^i\times3}$ as input, a novel feature structuring module FSNet  firstly  aggregates point-wise features encoded from ${P^i}$ into a 2D structured feature map $\sfm$ and feeds it to a point cloud decoding network to produce a coarse point cloud ${P^c}\in\mathbb{R}^{N^c\times3}$, where $N^i$ and $N^c$ denote the numbers of input points and points after coarse completion, respectively. Then, a sparse encoding module and a novel iterative feedback network IFNet are applied sequentially to upsample and refine the coarse completion ${P^c}$, followed by an offset regression module to obtain a dense point cloud ${P^d}\in\mathbb{R}^{N^d\times3}$ with fine-grained details, where $N^d$ denotes the number of points after dense completion.

\begin{figure*}
\centering
\includegraphics[width=\textwidth]{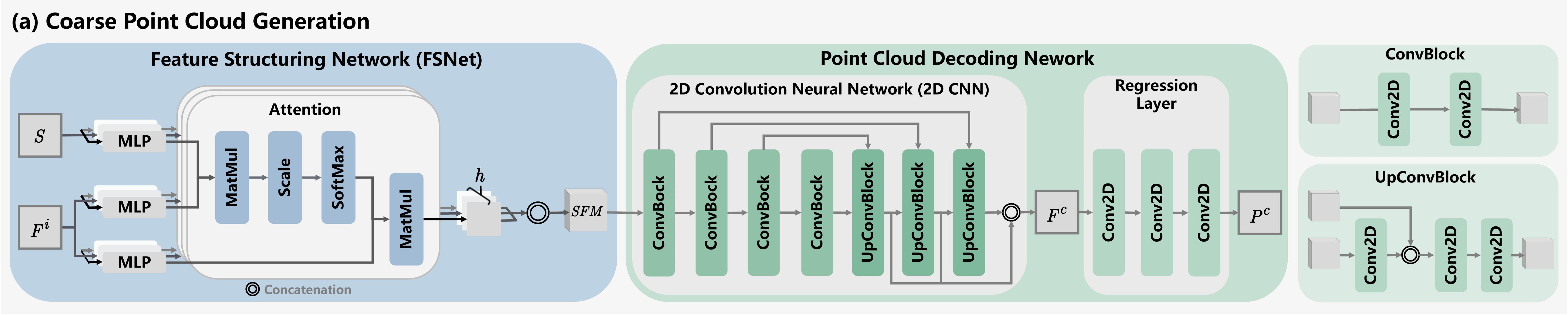}\\
\includegraphics[width=\textwidth]{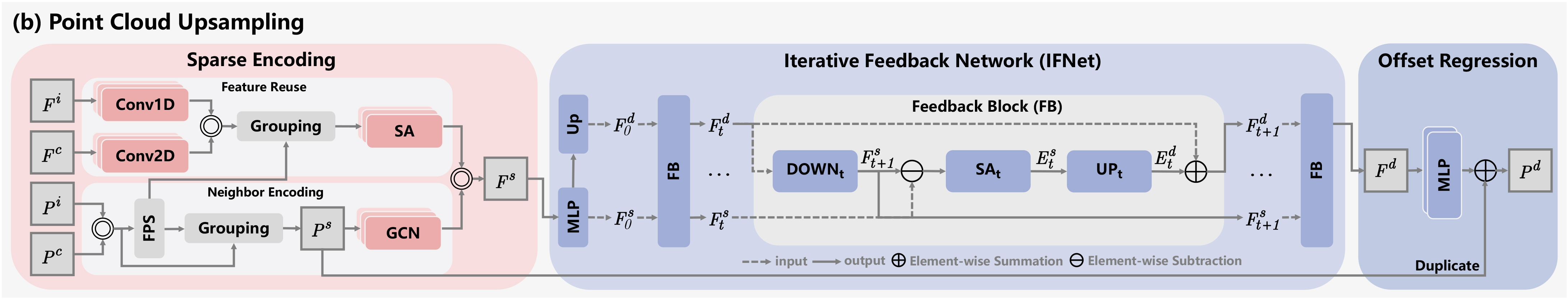}
\caption{Network architectures of (a) our coarse point cloud generation module and (b) our point cloud upsampling module.}
\label{fig:methods}
\end{figure*}

\subsection{Coarse Point Cloud Generation}
\subsubsection{Background}
The goal of the first stage is to generate a coarse and complete point cloud. The main challenge is to recover a coarse shape that has correct  global structures while maintaining sufficient local geometric details with respect to the input. To this end, we propose a feature structuring network, FSNet, that can adaptively aggregate point-wise features into a 2D feature map and then decode it into a coarse point cloud, as shown in Figure~\ref{fig:methods} (a). 

\begin{figure}
\centering
\includegraphics[width=\linewidth]{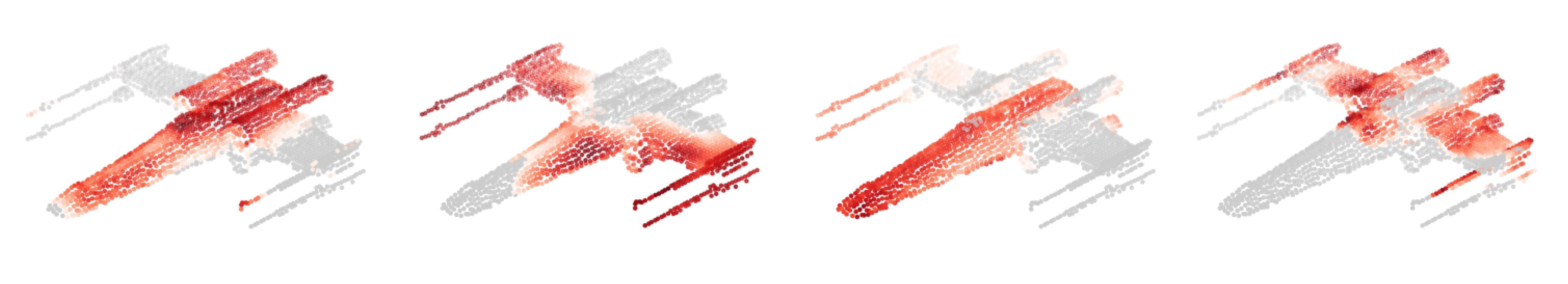}
\caption{Normalized attention weights of some representative channels in the $\sfm$.}
\label{fig:attention_weight_1}
\end{figure}

\subsubsection{Feature structuring network} 
To preserve permutation invariance of the input points, max pooling is widely used by previous methods, but it often leads to unavoidable information loss of local details. Inspired by Set Transformer~\cite{lee2019set}, we propose FSNet, which utilizes multi-head attention~\cite{vaswani2017attention} instead of max pooling to aggregate point-wise features in a data-driven way, while ignoring the order of the input points.

Specifically, given point-wise features  ${F^i}\in\mathbb{R}^{N^i\times C^i}$ extracted from the input point cloud, we explore the latent patterns in local regions via an attention mechanism. We first define a learnable set with $k$ vectors $S\in\mathbb{R}^{k\times C^i}$ and the ${F^i}$ as the queries and key-value pairs. Then, we linearly project the queries, keys and values $h$ times through different MLPs to dimension $d$. We further perform an attention function~\cite{vaswani2017attention} on the projected features, yielding $h$ matrices with shape $k\times d$, where each matrix is generated as the weighed sum of the projected input features and the attention weights which determine the geometric region corresponding to the channel extracted from the input point cloud.
These matrices are then concatenated to generate a 2D structured feature map $\sfm\in\mathbb{R}^{h \times k \times d}$ with shape $k \times d$ and $h$ channels.
Note that the output feature map is independent of the order of the input points, and naturally takes all vectors of input features into consideration thanks to the attention mechanism.

To better understand the learned $\sfm$, we visualize the normalized attention weights of some representative channels using heatmaps in Figure~\ref{fig:attention_weight_1}. We observe that the distribution of attention weights shows different patterns across channels; some  reveal semantic correlation, while others are geometrically complementary, indicating the structured feature map contains different combinations of local information from the input point cloud. 

\subsubsection{Point cloud decoding network}
With the learned $\sfm$, the next step is to generate a coarse but complete point cloud that  recovers the overall structure of the target object. The key challenge here is to infer the features of missing regions and recover the underlying structure for diverse topologies. 
Our key observation is that since $\sfm$ has a regular 2D format and contains rich local information about the input, we may model the point cloud structure via 2D convolution operations.
Thus, we build a decoding network that learns to generate a point cloud from the $\sfm$. It is composed of two parts: a 2D CNN with UNet structure and a regression layer. In more detail, given the $\sfm$ as input, the 2D CNN produces  coarse features ${F^c}\in\mathbb{R}^{C^c \times k \times d}$, and then the regression layer transforms ${F^c}$ to an intermediate output ${\bar{P}^c}$ with shape $ {3 \times (k/2) \times (d/2)} $, which is reshaped to form the corresponding coarse point cloud ${P^c\in \mathbb{R}^{N^c \times 3}}$. 

\begin{figure}
\centering\includegraphics[width=\linewidth]{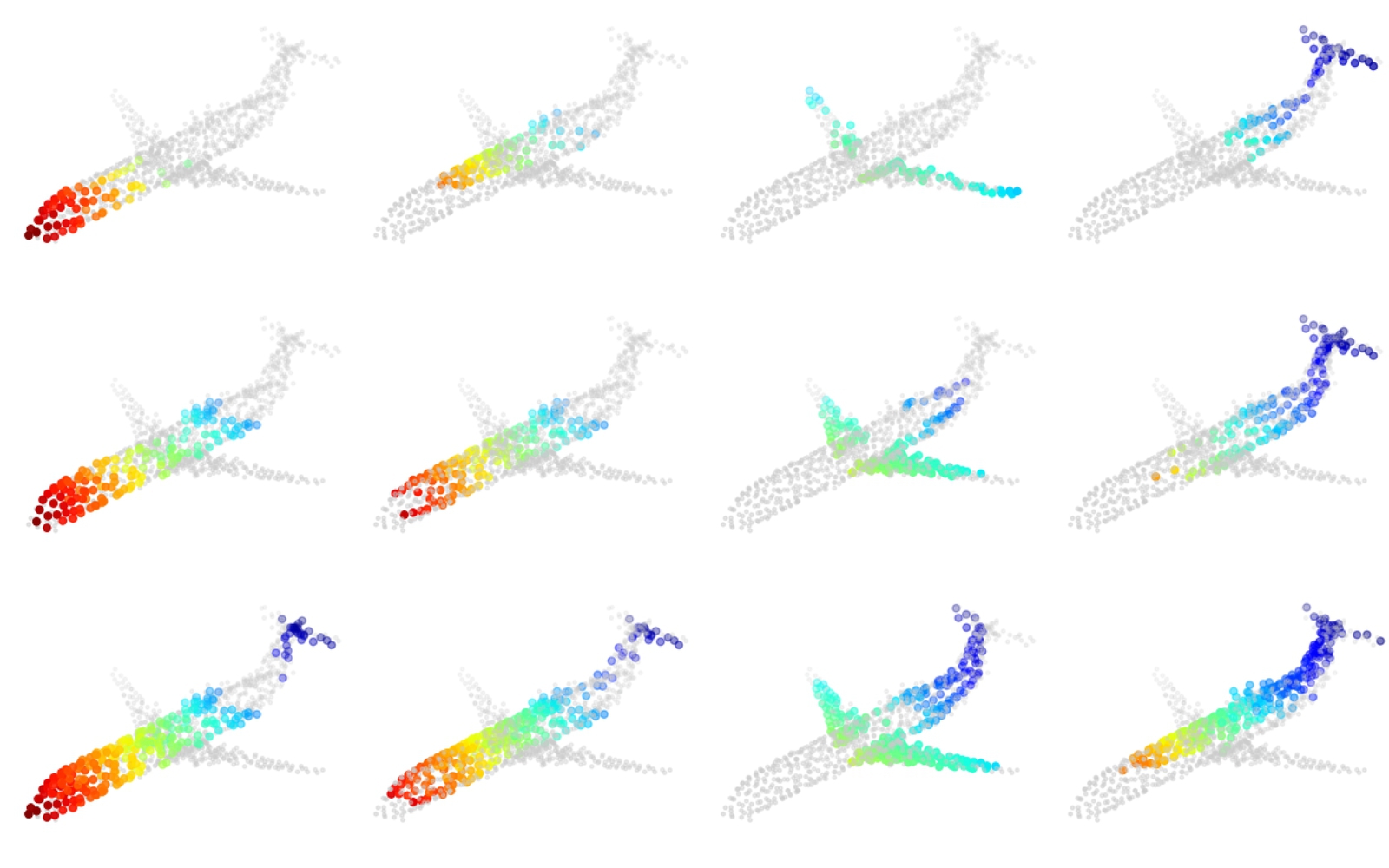}
\caption{Representative structures learned by the decoding network. Local patches in the 2D structured point cloud ${\bar{P}^c}$ correspond to local point regions, which  grow with  increasing patch size from the top row to the bottom row.
}
\label{fig:decoding_structure}
\end{figure}

Note that as the structure information is implicitly modeled in the 2D CNN, local patches in ${\bar{P}^c}$ actually correspond to local neighboring point regions, which represent the local structure of the object. Figure~\ref{fig:decoding_structure} shows some example local point regions corresponding to local patches in ${\bar{P}^c}$, with increasing sizes from the top  to bottom rows, where we use sliding windows of different sizes to extract the local patches.
We can see that point regions corresponding to smaller patches represent low-level structures of the object, and point regions grow with  increasing local patch size in ${\bar{P}^c}$ to represent  high-level structures of the object. This shows that the CNN-based decoding network can precisely model the structure of point cloud using the given $\sfm$.

\subsection{Point Cloud Upsampling}
\subsubsection{Approach}
After generating the coarse point cloud $P^c$, a point cloud upsampling stage  follows  to produce a dense point cloud ${P^d}$ with fine-grained details and uniform distribution. As Figure~\ref{fig:methods} (b) shows, our proposed  point cloud upsampling stage has three steps: sparse encoding, feature expansion, and offset regression.

\subsubsection{Sparse encoding}
The first step is a sparse encoding module which serves as sparse feature preparation and provides both local and contextual information to the following steps. It consists of two branches: a neighbor encoding path to preserve  geometric structures, and a feature reuse path to explore  structural details in local regions by revisiting the input features ${F^i}$ and coarse features ${F^c}$. For the neighbor encoding path, we first combine ${P^c}$ with ${P^i}$, followed by farthest point sampling and a grouping operation, which produces a sparse and relatively uniform point cloud ${P^s}\in\mathbb{R}^{N^s\times 3}$, where $N^s$ denotes the number of sparse sample points. We then follow the GCN structure in DGCNN~\cite{wang2019dynamic} with adaption of EdgeConv~\cite{wang2019dynamic} as the GCN layer, which sequentially aggregates features from the local neighbors for each point. For the feature reuse path, ${F^i}$ and ${F^c}$ are passed to a sequence of 1D and 2D convolution layers separately. The output features are concatenated, followed by a grouping operation and a sequence of self-attention units~\cite{zhang2019self} to enhance feature integration. At the end, features from the two paths are concatenated as the sparse features ${F^s}\in\mathbb{R}^{N^s\times C^s}$ and fed into the feature expansion step.

\begin{figure*}
\centering
\setlength\tabcolsep{0.6pt}
\newcommand\oneninthwidth{0.110}
\begin{tabular}{ccccccccc}
Input & NSFA & GRNet & PoinTr & VRCNet & PMP-Net & SnowflakeNet & Ours & GT \\
\includegraphics[width=\oneninthwidth\textwidth]{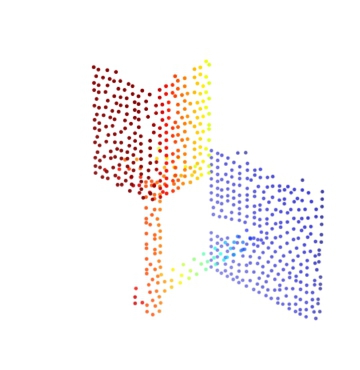} & 
\includegraphics[width=\oneninthwidth\textwidth]{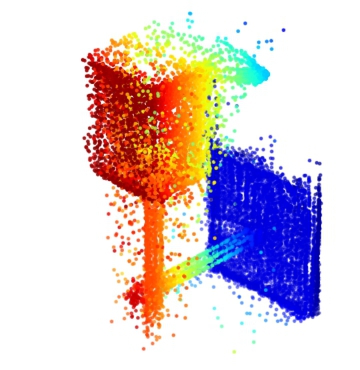} & 
\includegraphics[width=\oneninthwidth\textwidth]{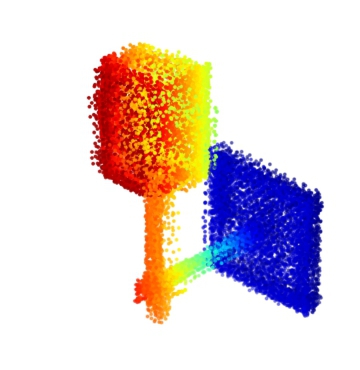} & 
\includegraphics[width=\oneninthwidth\textwidth]{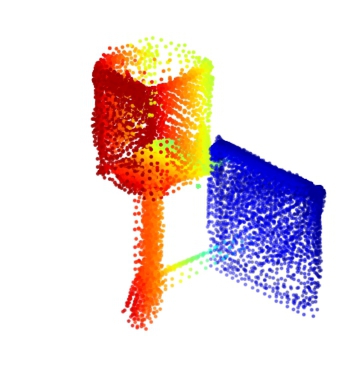} & 
\includegraphics[width=\oneninthwidth\textwidth]{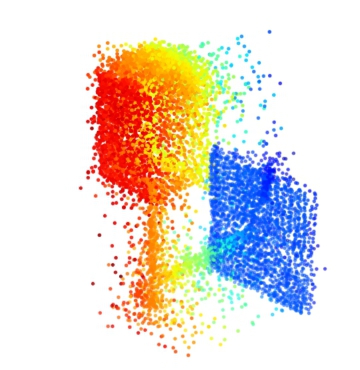} & 
\includegraphics[width=\oneninthwidth\textwidth]{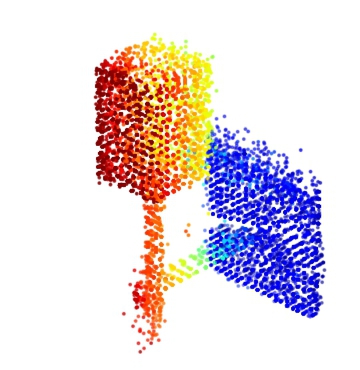} & 
\includegraphics[width=\oneninthwidth\textwidth]{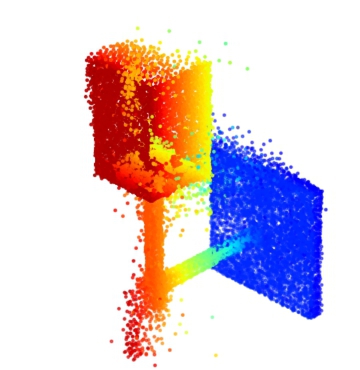} & 
\includegraphics[width=\oneninthwidth\textwidth]{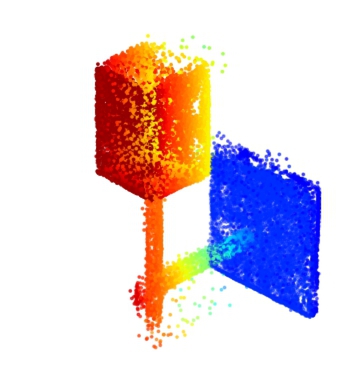} & 
\includegraphics[width=\oneninthwidth\textwidth]{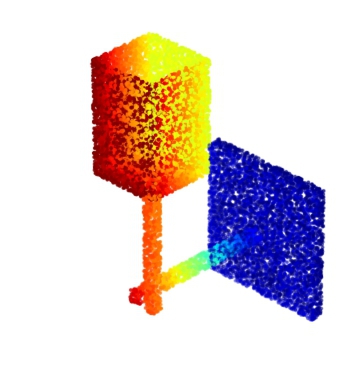}
\\
\includegraphics[width=\oneninthwidth\textwidth]{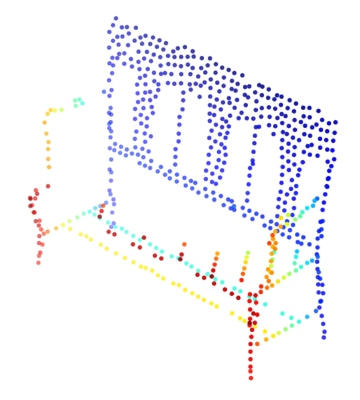} & 
\includegraphics[width=\oneninthwidth\textwidth]{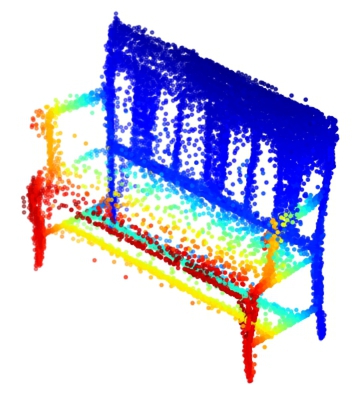} & 
\includegraphics[width=\oneninthwidth\textwidth]{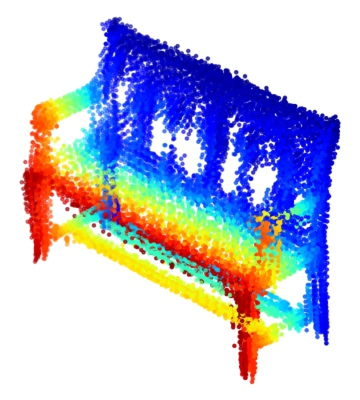} & 
\includegraphics[width=\oneninthwidth\textwidth]{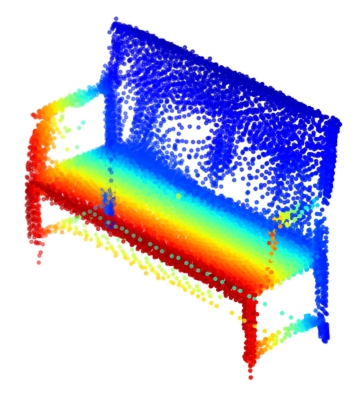} & 
\includegraphics[width=\oneninthwidth\textwidth]{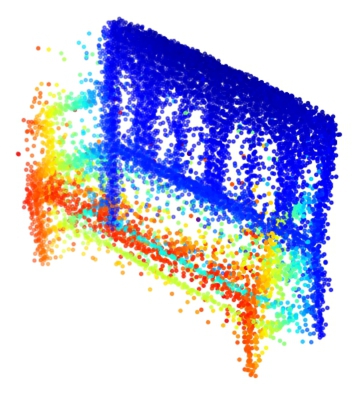} & 
\includegraphics[width=\oneninthwidth\textwidth]{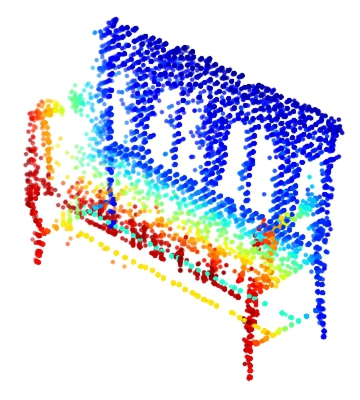} & 
\includegraphics[width=\oneninthwidth\textwidth]{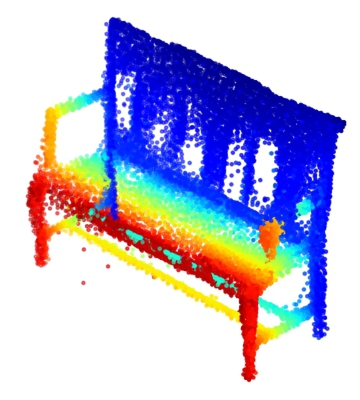} & 
\includegraphics[width=\oneninthwidth\textwidth]{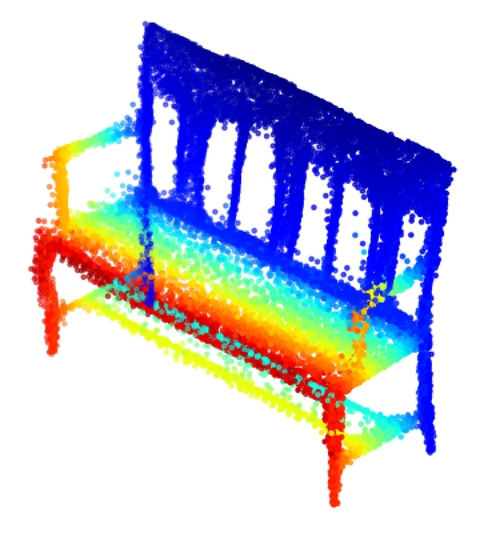} & 
\includegraphics[width=\oneninthwidth\textwidth]{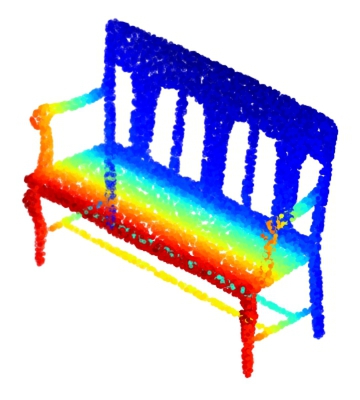}
\\
\includegraphics[width=\oneninthwidth\textwidth]{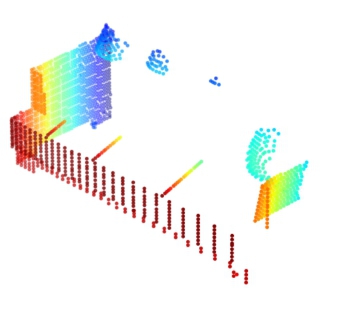} & 
\includegraphics[width=\oneninthwidth\textwidth]{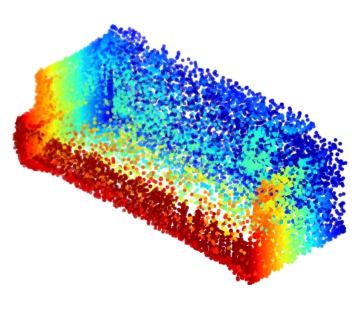} & 
\includegraphics[width=\oneninthwidth\textwidth]{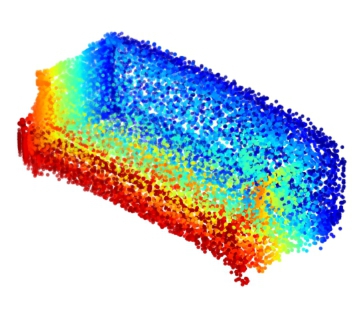} & 
\includegraphics[width=\oneninthwidth\textwidth]{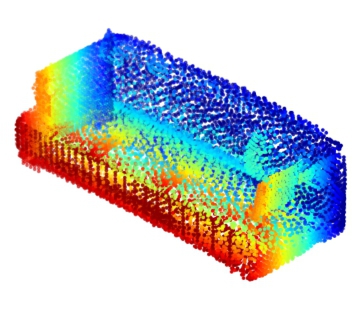} & 
\includegraphics[width=\oneninthwidth\textwidth]{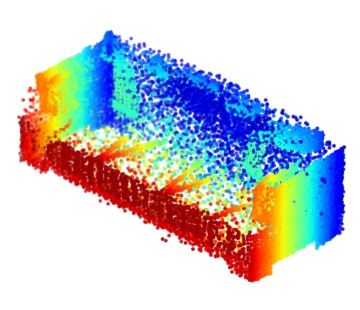} & 
\includegraphics[width=\oneninthwidth\textwidth]{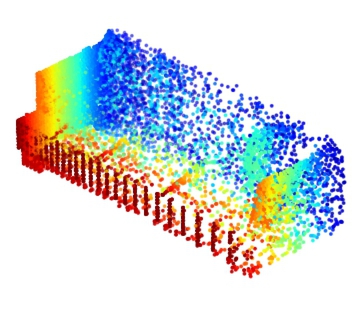} & 
\includegraphics[width=\oneninthwidth\textwidth]{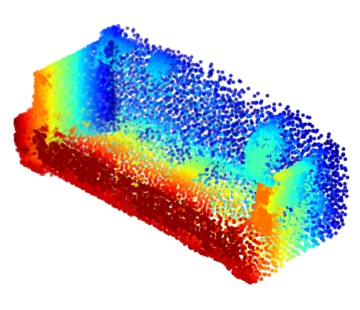} & 
\includegraphics[width=\oneninthwidth\textwidth]{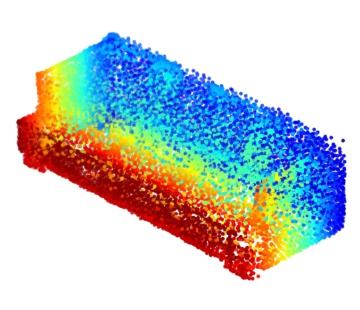} & 
\includegraphics[width=\oneninthwidth\textwidth]{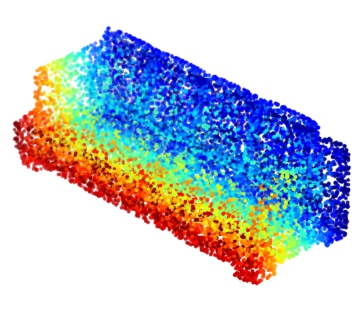}
\\
\includegraphics[width=\oneninthwidth\textwidth]{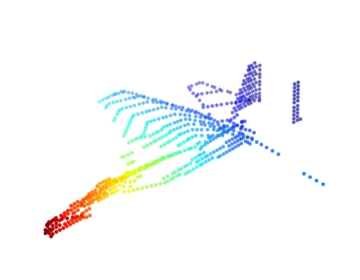} & 
\includegraphics[width=\oneninthwidth\textwidth]{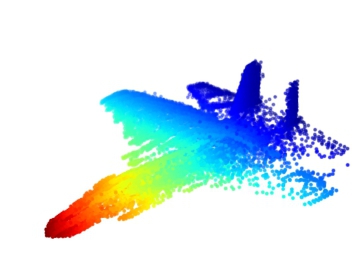} & 
\includegraphics[width=\oneninthwidth\textwidth]{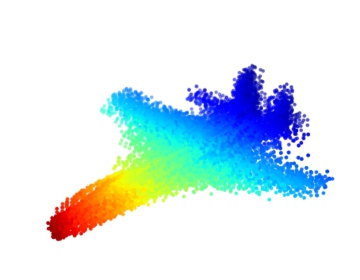} & 
\includegraphics[width=\oneninthwidth\textwidth]{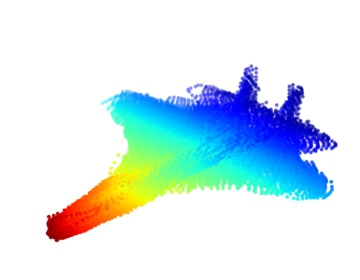} & 
\includegraphics[width=\oneninthwidth\textwidth]{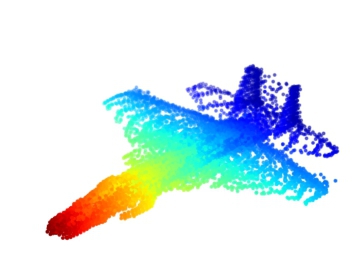} & 
\includegraphics[width=\oneninthwidth\textwidth]{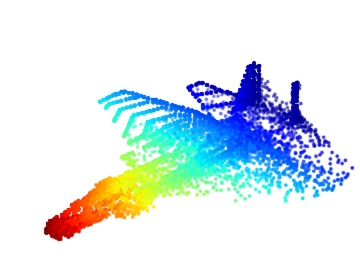} & 
\includegraphics[width=\oneninthwidth\textwidth]{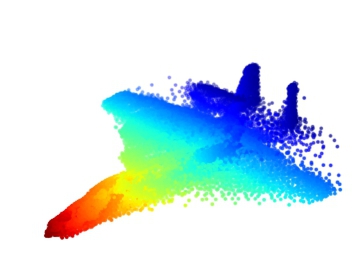} & 
\includegraphics[width=\oneninthwidth\textwidth]{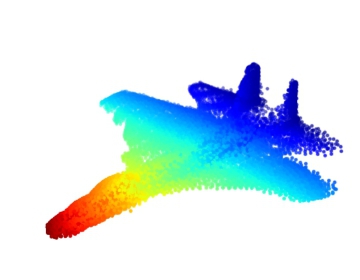} & 
\includegraphics[width=\oneninthwidth\textwidth]{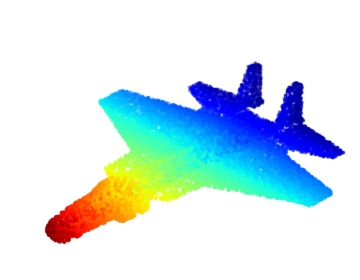}
\end{tabular}
\caption{Completion results on the ShapeNet dataset.}
\label{fig5}
\end{figure*}

\subsubsection{Iterative feedback network}
Inspired by DBPN~\cite{haris2018deep} for image super resolution tasks, we introduce a feedback mechanism into feature expansion by constructing an iterative feedback network (IFNet) that can effectively expand the sparse features through multi-step refinement. The network first reduces the dimension of $F^s$ from $C^s$ to $c$ using an MLPs and expands it through an upsampling unit, producing the initial sparse features ${F^s_0}\in\mathbb{R}^{N^s\times c}$ and dense point features ${F^d_0}\in\mathbb{R}^{rN^s\times c}$, where $r$ denotes the upsampling ratio. It then feeds the initial features into a sequence of feedback blocks (FBs) to perform self-correction. 

More specifically, a feedback block comprises an upsampling unit, a downsampling unit and a self-attention unit,  defined as: 
\begin{align}
{\rm scale\ down} &: {F^s_{t+1}} = {\rm DOWN_t}({F^d_t},r) \\
{\rm difference} &: {E^s_t} = {\rm SA_t}({F^s_{t+1}} - {F^s_t}) \\
{\rm scale\ difference\ up} &: {E^d_t} = {\rm UP_t}({E^s_t},r) \\
{\rm corrected\ features} &: {F^d_{t+1}} = {F^d_t} + {E^d_t}
\end{align}
where ${\rm SA}$ denotes the self-attention unit~\cite{zhang2019self}, and ${\rm UP}(F,r)$ and ${\rm Down}(F,r)$ denote upsampling and downsampling operations on features $F$ with a ratio $r$. Detailed network architectures are shown in the supplementary materials.

Each feedback block takes sparse features ${F^s_{t}}\in\mathbb{R}^{N^s\times c}$ and dense features ${F^d_{t}}\in\mathbb{R}^{rN^s\times c}$ as inputs, and maps the dense features ${F^d_{t}}$ to  new sparse features ${F^s_{t+1}}\in\mathbb{R}^{N^s\times c}$. The difference ${E^s_t}\in\mathbb{R}^{N^s\times c}$ between known sparse features ${F^s_{t}}$ and the reconstructed sparse features ${F^s_{t+1}}$ is computed, followed by a self-attention unit to produce a more discriminative error by capturing long-range
contextual information. Afterwards, the difference is mapped to  intermediate dense features ${E^d_t}\in\mathbb{R}^{rN^s\times c}$, and the corrected dense features ${F^d_{t+1}}\in\mathbb{R}^{rN^s\times c}$ are obtained by adding ${E^d_t}$ to ${F^d_{t}}$. Intuitively, each block performs a self-correction procedure by feeding back the projection error to the initial dense features, which enables the networks to produce dense features that help to produce a dense point cloud with fine-grained details and uniform distribution. 

\subsubsection{Offset regression}
In the last process, we generate the dense point cloud ${P^d}$ based on the produced dense features ${F^d}\in\mathbb{R}^{N^d\times c}$ by predicting offsets between the coordinates of points in the sparse and dense point clouds, where the residual offsets are regressed through two layers of MLPs and the sparse point cloud ${P^s}$ is replicated $r$ times before residual summation. Note that the number of dense completion points ${N^d} = r{N^s}$.

\subsection{Loss function}
We choose Chamfer distance (CD) as the reconstruction loss due to its efficiency, following PSGN~\cite{fan2017point} and TopNet~\cite{tchapmi2019topnet}. The Chamfer distance between two point sets ${X}$ and ${Y}$ is defined as:
\begin{equation}
    L_{CD}\left (X, Y \right ) = L_{X,Y} + L_{Y,X},
\end{equation}
where
\begin{equation}
    L_{X, Y}=\frac{1}{\left | X \right | } \sum_{x\in X} \min_{y \in Y} \left | \left | x-y \right |  \right |_2^{2}.
\end{equation}
Since we generate the complete point cloud in a coarse-to-fine fashion, we jointly optimize the coarse point cloud $P^c$ and dense point cloud $P^d$ via the CD loss. The overall training loss is defined as:
\begin{equation}
    L=L_{CD}\left ( P^s, P^g\right )+L_{CD}\left ( P^d, P^g\right )
\end{equation}
where ${P^g}\in\mathbb{R}^{N^g\times 3}$ denotes the ground truth point cloud, and $N^g$ denotes the number of ground truth points.

\section{Experiments}

\subsection{Datasets}
We evaluated our method using three popular benchmarks, including ShapeNet~\cite{wu20153d}, MVP~\cite{pan2021variational}, and KITTI~\cite{geiger2013vision}:

\subsubsection{ShapeNet}
The ShapeNet dataset for point cloud completion is derived from PCN~\cite{yuan2018pcn}; 30,974 samples are selected in 8 categories. The ground truth point clouds containing 16,384 points are uniformly sampled from mesh surfaces. The partial point clouds are generated by back-projecting 2.5D depth maps from 8 random views into 3D. For a fair comparison, we use the same training / validation / test splits as PCN.

\subsubsection{MVP}
The MVP dataset consists of 16 categories of partial and complete point clouds generated from CAD-models selected from the ShapeNet dataset~\cite{wu20153d}. There are 62400 and 41600 shape pairs in the training and testing sets, respectively. Unlike other datasets, the complete point clouds in MVP have different resolutions, including 2048, 4096, 8192 and 16384, so can be used to evaluate completion quality at different resolutions. We use the same training / test splits as in VRCNet~\cite{pan2021variational} to evaluate our method.

\subsubsection{KITTI}
The KITTI dataset is composed of a sequence of real-world LiDAR scans, also derived from PCN~\cite{yuan2018pcn}. For each frame, a car point cloud is extracted within the car object bounding boxes, resulting in 2,401 partial point clouds. There is no ground truth for this dataset.

\begin{table*}
\centering
\begin{tabular}{l|rrrrrrrr|r}
\toprule
\multicolumn{1}{c|}{Method} & Plane      & Cabinet       & Car           & Chair         & Lamp          & Sofa          & Table         & Vessel    & Avg.       \\ \midrule \midrule
PCN~\cite{yuan2018pcn}                         & 1.40          & 4.45          & 2.45          & 4.84          & 6.24          & 5.13          & 3.57          & 4.06          & 4.02          \\
TopNet~\cite{tchapmi2019topnet}                      & 2.15          & 5.62          & 3.51          & 6.35          & 7.50          & 6.95          & 4.78          & 4.36          & 5.15          \\
MSN~\cite{liu2020morphing}                         & 1.54          & 7.25          & 4.71          & 4.54          & 6.48          & 5.89          & 3.80          & 3.85          & 4.76          \\
NSFA~\cite{zhang2020detail}                        & 1.75          & 5.31          & 3.43          & 5.01          & 4.73          & 6.41          & 4.00          & 3.56          & 4.27          \\
PF-Net~\cite{huang2020pf}                      & 1.55          & 4.43          & 3.12          & 3.96          & 4.21          & 5.87          & 3.35          & 3.89          & 3.80          \\
CRN~\cite{wang2020cascaded}                         & 1.46          & 4.21          & 2.97          & 3.24          & 5.16          & 5.01          & 3.99          & 3.96          & 3.75          \\
GRNet~\cite{xie2020grnet}                       & 1.53          & 3.62          & 2.75          & 2.95          & 2.65          & 3.61          & 2.55          & 2.12          & 2.72          \\
PoinTr~\cite{yu2021pointr}                      & 0.99          & 4.80          & 2.52          & 3.68          & 3.07          & 6.53          & 3.10          & 2.02          & 3.34
   \\
VRCNet~\cite{pan2021variational}                       & 1.53          & 4.66          & 2.66          & 4.62          & 5.50          & 5.70          & 4.39          & 3.58          & 4.08          \\
PMP-Net~\cite{wen2021pmp}                      & 1.22	         & 4.18	         & 2.85	         & 3.51	         & 2.14	         & 4.22	         & 2.89	         & 1.88	         & 2.86          \\
SnowflakeNet~\cite{xiang2021snowflakenet}                 & 0.86          & \textbf{3.40} & 2.36          & 2.68          & 2.06          & 4.46          & 2.16          & \textbf{1.74} & 2.47          \\
ASHF-Net~\cite{zong2021ashf}                    & 1.40          & 3.49          & 2.32          & 2.82          & 2.52          & \textbf{3.48} & 2.42          & 1.99          & 2.55          \\ \midrule
Ours                         & \textbf{0.80} & 3.53          & \textbf{2.13} & \textbf{2.48}          & \textbf{1.79} & 3.64          & \textbf{1.98} & 1.81          & \textbf{2.27} \\ 
\bottomrule
\end{tabular}
\caption{Shape completion results (CD loss$\times 10^4$) on the ShapeNet dataset.}
\label{tab:ShapeNet}
\end{table*}

\begin{table*}[t!]
\centering
\begin{tabular}{l|rrrrrrrr}
\toprule
\multicolumn{1}{c|}{\multirow{2}{*}{Points}} & \multicolumn{2}{c|}{2,048}                           & \multicolumn{2}{c|}{4,096}                           & \multicolumn{2}{c|}{8,192}                           & \multicolumn{2}{c}{16,384}     \\ \cline{2-9} 
\multicolumn{1}{c|}{}                        & CD            & \multicolumn{1}{c|}{F1}             & CD            & \multicolumn{1}{c|}{F1}             & CD            & \multicolumn{1}{c|}{F1}             & CD            & F1             \\ \midrule \midrule
PCN~\cite{yuan2018pcn}                                         & 9.77          & \multicolumn{1}{c|}{0.320}          & 7.96          & \multicolumn{1}{c|}{0.458}          & 6.99          & \multicolumn{1}{c|}{0.563}          & 6.02          & 0.638          \\
TopNet~\cite{tchapmi2019topnet}                                       & 10.11         & \multicolumn{1}{c|}{0.308}          & 8.20          & \multicolumn{1}{c|}{0.440}          & 7.00          & \multicolumn{1}{c|}{0.533}          & 6.36          & 0.601          \\
MSN~\cite{liu2020morphing}                                          & 7.90          & \multicolumn{1}{c|}{0.432}          & 6.17          & \multicolumn{1}{c|}{0.585}          & 5.42          & \multicolumn{1}{c|}{0.659}          & 4.90          & 0.710          \\
CRN~\cite{wang2020cascaded}                                         & 7.25          & \multicolumn{1}{c|}{0.434}          & 5.83          & \multicolumn{1}{c|}{0.569}          & 4.90          & \multicolumn{1}{c|}{0.680}          & 4.30          & 0.740          \\
ECG~\cite{pan2020ecg}                                         & 6.64          & \multicolumn{1}{c|}{0.476}          & 5.41          & \multicolumn{1}{c|}{0.585}          & 4.18          & \multicolumn{1}{c|}{0.690}          & 3.58          & 0.753          \\
VRCNet~\cite{pan2021variational}                                      & 5.96          & \multicolumn{1}{c|}{0.499}          & 4.70          & \multicolumn{1}{c|}{0.636}          & 3.64          & \multicolumn{1}{c|}{0.727}          & 3.12          & 0.791          \\ 
PMPNet~\cite{wen2021pmp}                                      & 6.33          & \multicolumn{1}{c|}{0.479}          & 4.63          & \multicolumn{1}{c|}{0.584}          & 3.52          & \multicolumn{1}{c|}{0.680}          & 2.79          & 0.753        \\
SnowflakeNet~\cite{xiang2021snowflakenet}                                 & 6.06          & \multicolumn{1}{c|}{0.500} & 4.80     & \multicolumn{1}{c|}{0.615}               & 3.49          & \multicolumn{1}{c|}{0.739}          & 2.75          & 0.805          \\ \midrule
Ours                                         & \textbf{5.53} & \multicolumn{1}{c|}{\textbf{0.503}} & \textbf{4.20} & \multicolumn{1}{c|}{\textbf{0.648}} & \textbf{3.19} & \multicolumn{1}{c|}{\textbf{0.760}} & \textbf{2.33} & \textbf{0.810} \\ 
\bottomrule
\end{tabular}
\caption{Shape completion results (CD loss$\times10^4$ and F-score@1\%) with various resolutions on the MVP dataset.}
\label{tab:MVP}
\end{table*}

\subsection{Implementation Details}
\label{sec:implementation_details}
We use DGCNN~\cite{wang2019dynamic} as the encoder to extract features from the input point cloud. Both $d$ and $k$ are set to $64$ for generating a coarse completion ${P^c}$ with ${N^c}=1024$ points. The number of channels $h$ for $\sfm$ is set to $32$, determined by experiment. We typically sample ${N^s}=1024$ points to obtain the sparse point cloud ${P^s}$, which is sufficient to cover the overall structures of the objects. We may generate dense output $P^d$ with various resolutions, including ${N^d} = 2048, 4096, 8192, 16384$, where the upsampling ratio $r$ is set to 2, 4, 8, and 16, respectively.

Our networks are implemented using PyTorch and optimized using an Adam optimizer~\cite{kingma2014adam} with $\beta_1=0.9$ and $\beta_2=0.999$. The initial learning rate is $10^{-3}$ and is multiplied by $0.7$ per $10$ epochs. We train the networks with a batch size of 32 on four NVIDIA TITAN Xp GPUs.

\subsection{Comparison with State-of-the-Art}
We have compared our method to several  state-of-the-arts point cloud completion approaches,  both quantitatively and visually. 

\subsubsection{Results on ShapeNet}
A quantitative comparison of our method to other  methods on the ShapeNet dataset is shown in Table~\ref{tab:ShapeNet}: our method outperforms all competitive methods in terms of average CD across all categories.  Qualitative results are shown in Figure~\ref{fig5}: our method can precisely predict the missing parts of the input while other methods tend to output blurred point clouds in the missing region. Moreover, our method can produce more uniformly-distributed dense point clouds with less noise than other methods, benefiting from IFNet with its self-correction procedure. 

\begin{table*}[t!]
\renewcommand{\arraystretch}{1.05}
\centering
\begin{tabular}{l|rrrrrr}
\toprule
\multirow{2}{*}{Methods} & FD               & MMD              & Consistency                                     & \multicolumn{3}{c}{Uniformity}                   \\ \cline{5-7} 
                        & ($\times10^{-3}$) & ($\times10^{-3}$) & ($\times10^{-3}$) & 0.4\%          & 0.8\%          & 1.2\%          \\ \midrule \midrule
PCN~\cite{yuan2018pcn}                    & 2.235            & 1.366            & 1.557                                          & 3.662          & 7.710          & 10.823         \\
TopNet~\cite{tchapmi2019topnet}                  & 5.354            & 0.636            & 0.568                                          & 1.353          & 1.219          & 0.950          \\
MSN~\cite{liu2020morphing}                    & 0.434   & 2.259            & 1.951                                          & 0.822          & 0.523          & 0.383          \\
NSFA~\cite{zhang2020detail}                   & 1.281            & 0.891            & 0.491                                          & 0.992          & 0.767          & 0.552          \\
PF-Net~\cite{huang2020pf}                 & 1.137            & 0.792            & 0.436                                          & 0.881          & 0.682          & 0.491          \\
CRN~\cite{wang2020cascaded}                    & 1.023            & 0.872            & 0.431                                          & 0.870          & 0.673          & 0.485          \\
GRNet~\cite{xie2020grnet}                   & 0.816            & 0.568            & 0.313                                          & 0.632          & 0.489          & 0.352          \\
VRCNet~\cite{pan2021variational}                 & 2.586            & 0.378            & \textbf{0.259}                                 & 0.751          & 0.582          & 0.416          \\
SnowflakeNet~\cite{xiang2021snowflakenet}                  & \textbf{0.220}            & 0.664            & 0.557                                           & 0.979          & 0.730          & 0.499          \\
ASHF-Net~\cite{zong2021ashf}        & 0.773            & 0.541            & 0.298                                 & 0.602          & 0.466          & 0.335          \\ \midrule
Ours                 & 0.586            & \textbf{0.336}        & 0.314                                          & \textbf{0.506} & \textbf{0.313} & \textbf{0.148} \\ 
\bottomrule
\end{tabular}
\caption{Shape completion results on the KITTI dataset: fidelity distance (FD), minimal matching distance (MMD), consistency, and uniformity.}
\label{tab:KITTI}
\end{table*}

\begin{figure*}[t!]
  \centering
  \begin{tabular}{ c c c c c c }
    Input & GRNet & SnowflakeNet & VRCNet & Ours
    \\
    \multirow{2}{*}{\begin{minipage}{.1\textwidth}
      \includegraphics[width=\linewidth]{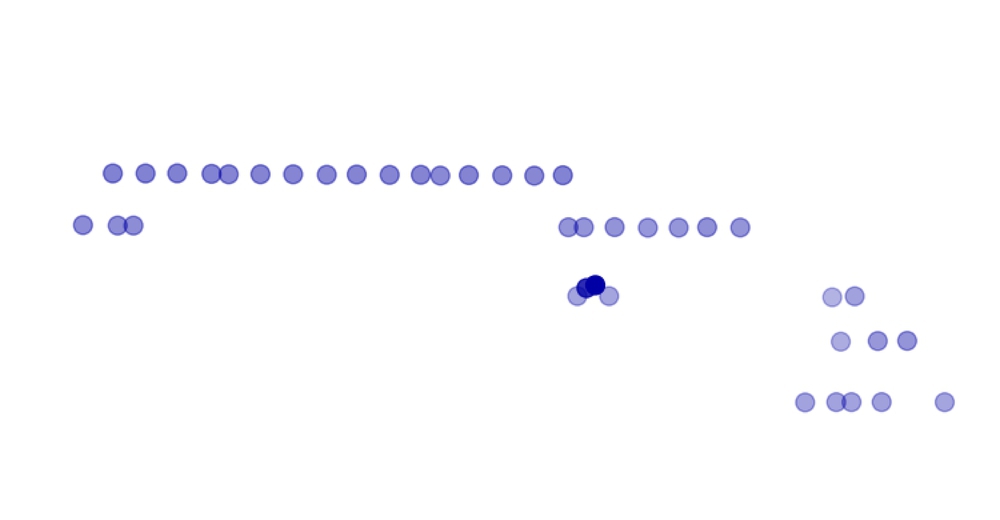}
    \end{minipage}}
    &
    \begin{minipage}{.18\textwidth}
      \includegraphics[width=\linewidth]{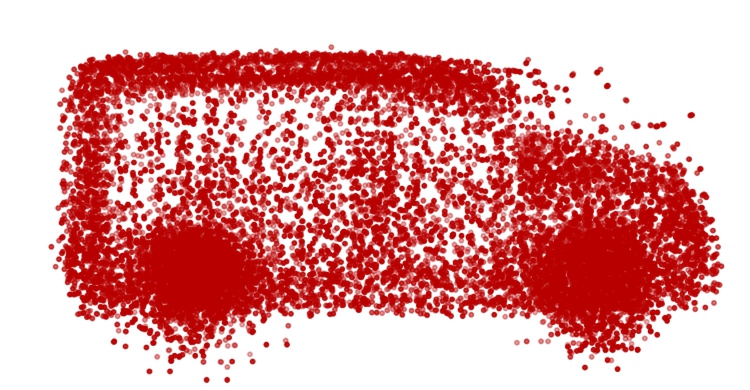}
    \end{minipage}
    &
    \begin{minipage}{.18\textwidth}
      \includegraphics[width=\linewidth]{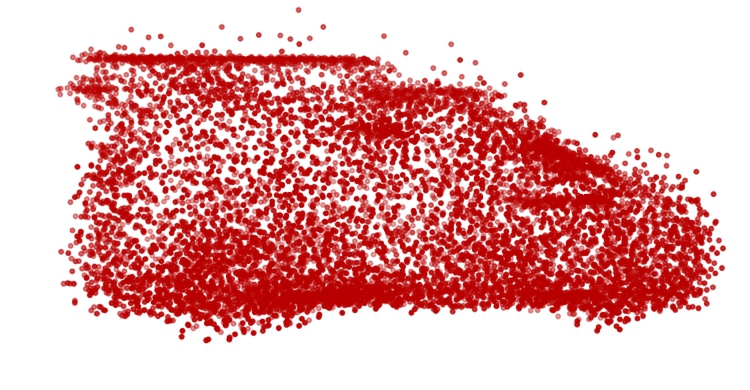}
    \end{minipage}
    & 
    \begin{minipage}{.18\textwidth}
      \includegraphics[width=\linewidth]{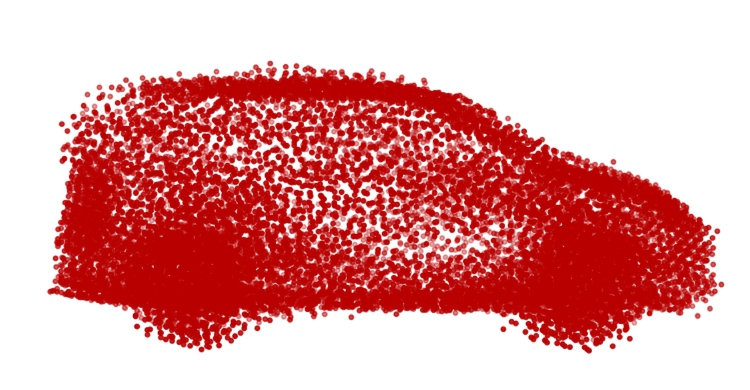}
    \end{minipage}
    &
    \begin{minipage}{.18\textwidth}
      \includegraphics[width=\linewidth]{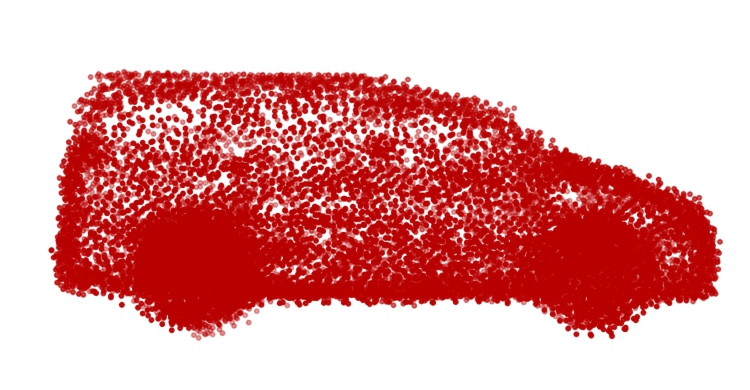}
    \end{minipage}
    \\ 
    &
    \begin{minipage}{.18\textwidth}
      \includegraphics[width=\linewidth]{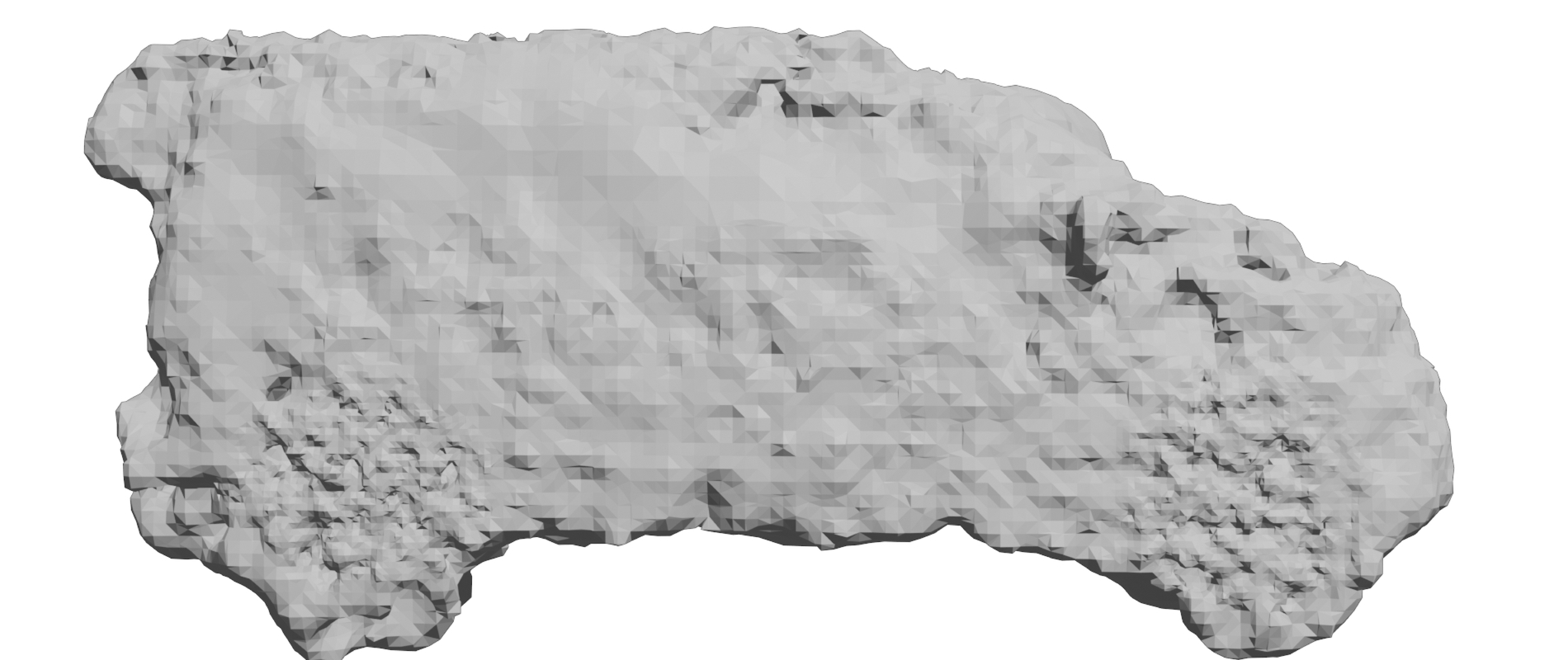}
    \end{minipage}
    &
    \begin{minipage}{.18\textwidth}
      \includegraphics[width=\linewidth]{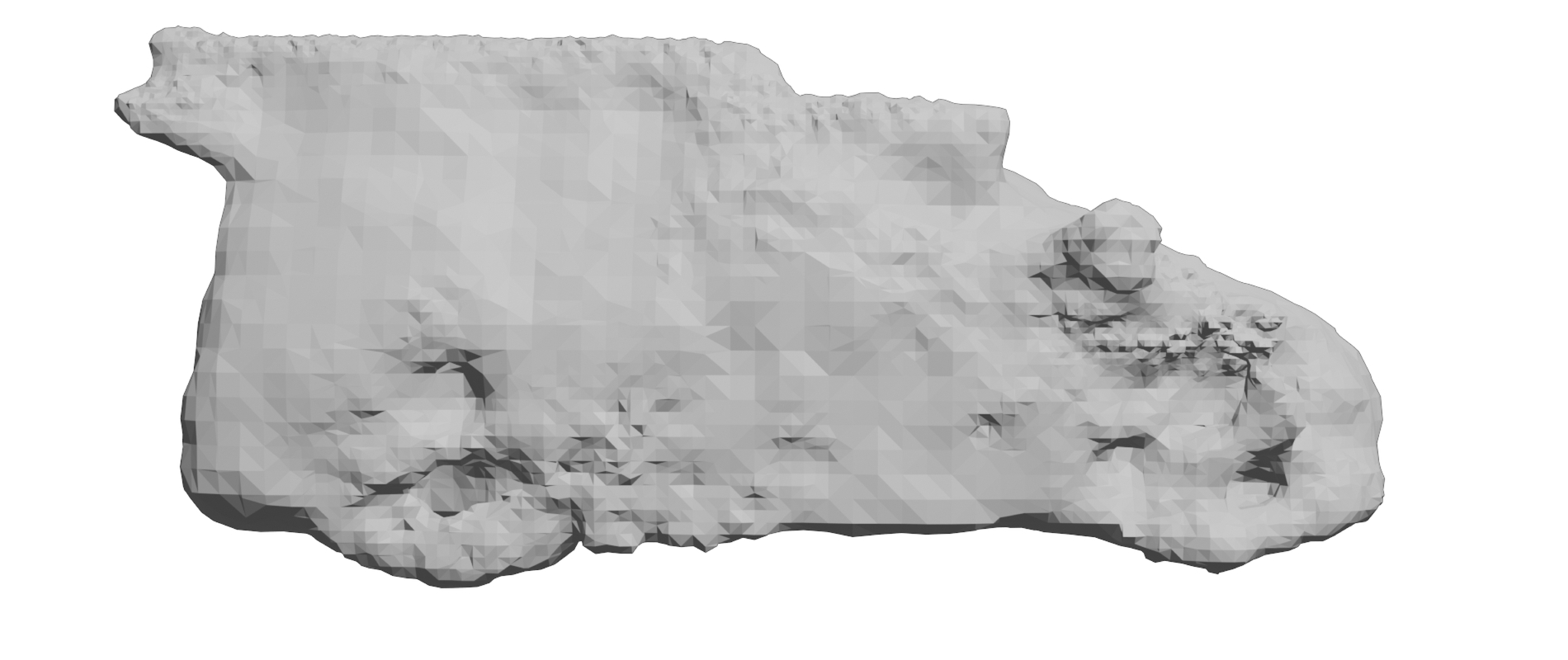}
    \end{minipage}
    & 
    \begin{minipage}{.18\textwidth}
      \includegraphics[width=\linewidth]{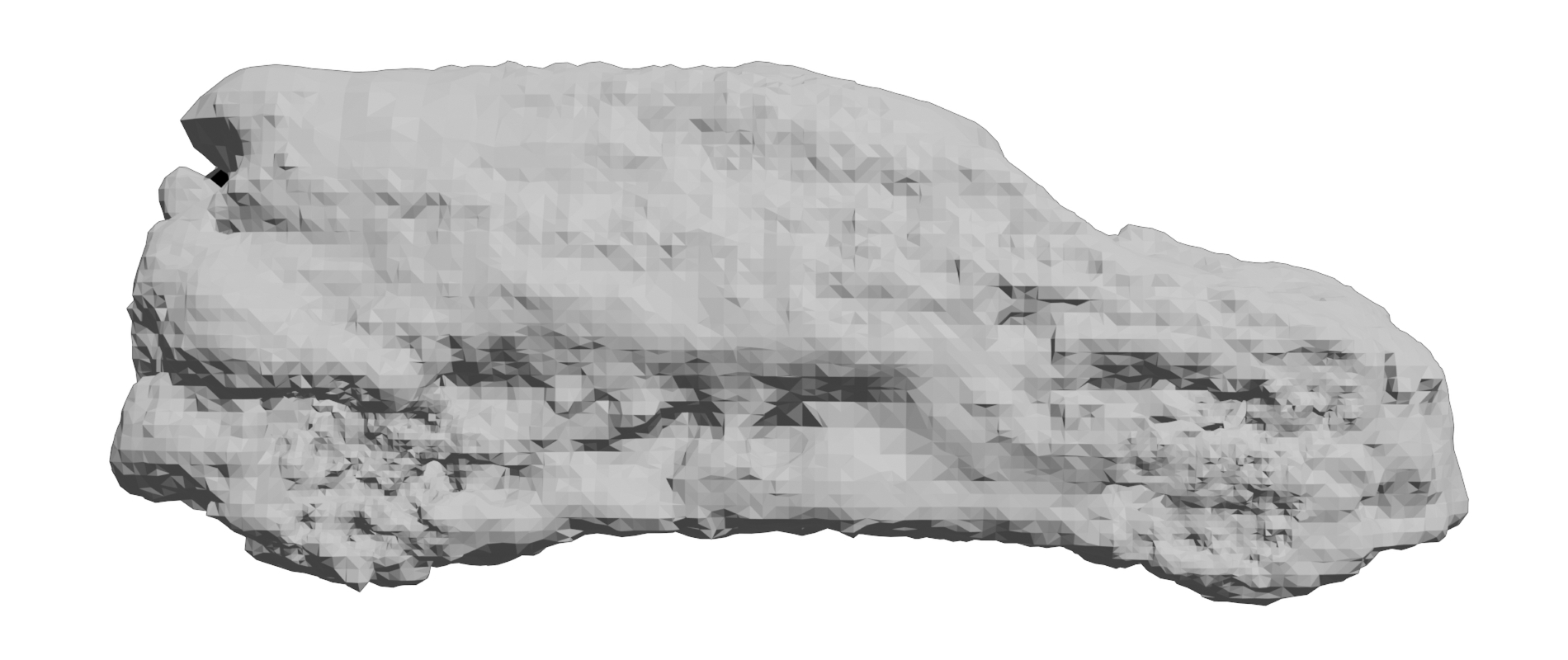}
    \end{minipage}
    &
    \begin{minipage}{.18\textwidth}
      \includegraphics[width=\linewidth]{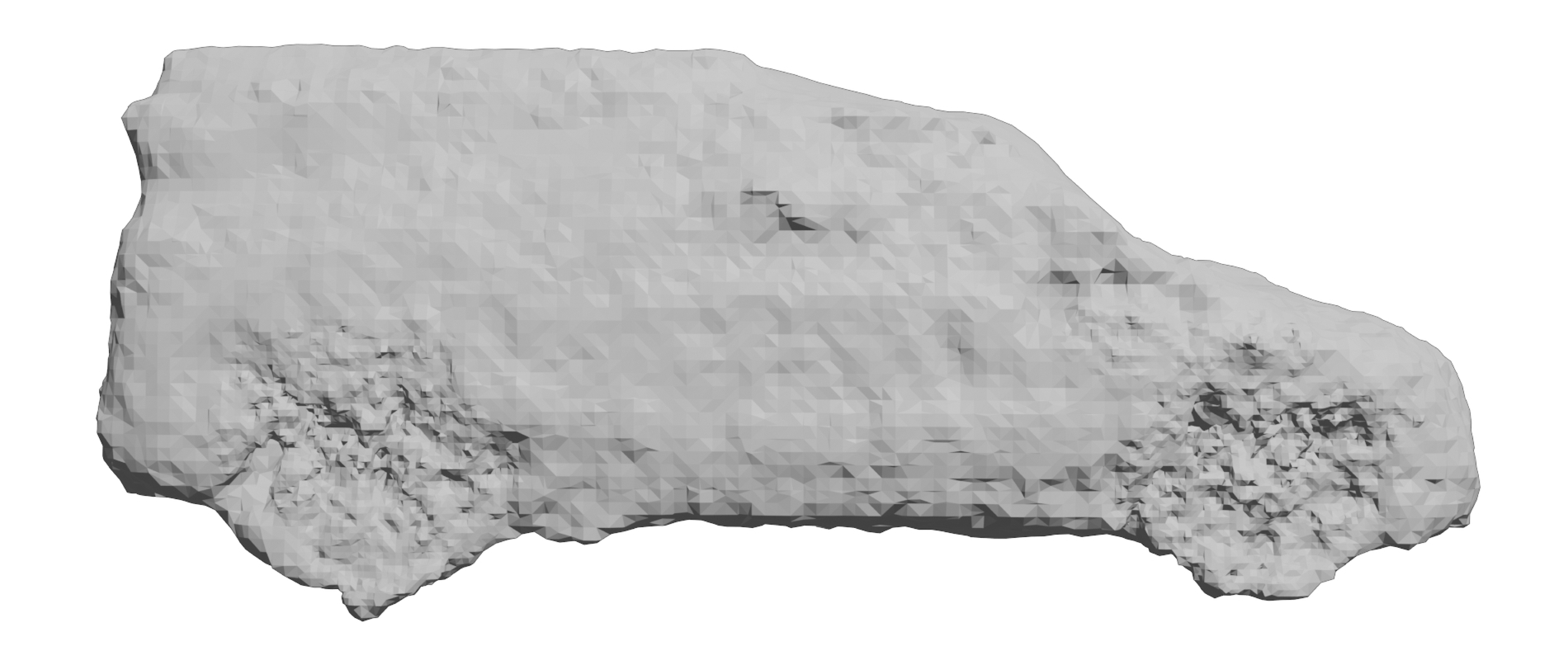}
    \end{minipage}
  \end{tabular}
  \includegraphics[width=0.48\linewidth]{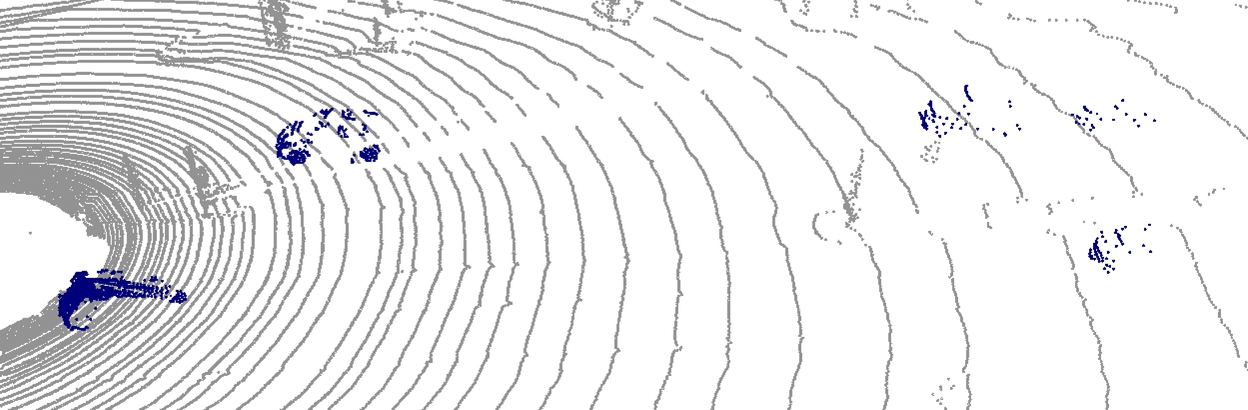}
  \includegraphics[width=0.48\linewidth]{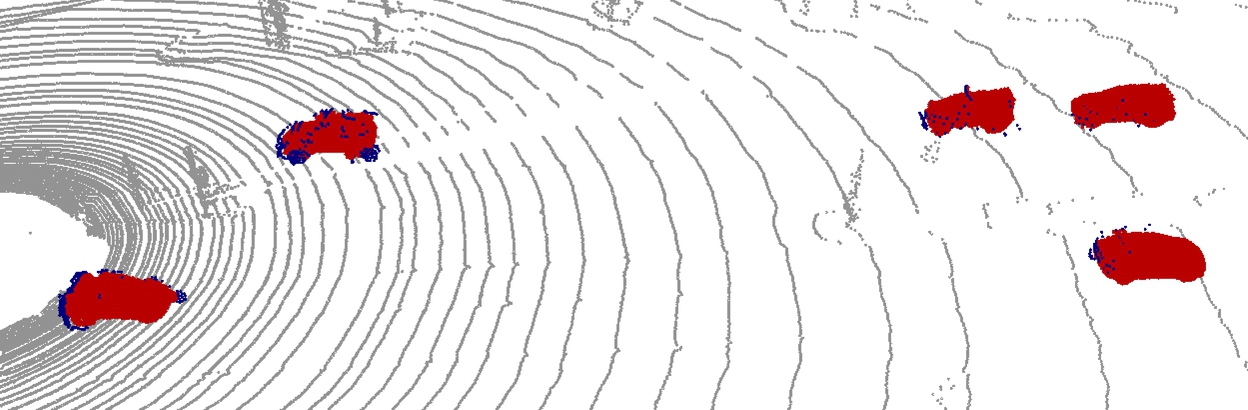}
  \caption{Qualitative point cloud completion and surface reconstruction results on the KITTI dataset.}
  \label{fig:KITTI}
\end{figure*}

\subsubsection{Results on MVP}
Since our method can generate complete point cloud with various resolutions by modifying the upsampling rate $r$, we compare our method with existing methods that support multi-resolution completion. Except for CD loss, we also use F-Score~\cite{knapitsch2017tanks} to evaluate the distance between objects as in~\cite{pan2021variational}. As shown in Table~\ref{tab:MVP}, our method outperforms all the other methods in terms of both CD and F-Score@1\%. 

\subsubsection{Results on KITTI}
Since there are no complete ground truth point clouds for KITTI, we follow the protocol of ASHF-Net~\cite{zong2021ashf} to evaluate the performance, where Fidelity, Minimal Matching Distance (MMD), Consistency and Uniformity are used as evaluation metrics. Table~\ref{tab:KITTI} shows the advantages of the point clouds generated by our method over the other methods in terms of distribution uniformity, which demonstrates that the proposed IFNet is beneficial to produce uniformly distributed point cloud through the iterative self-correction procedure. Besides, our approach achieves the lowest MMD among all the methods.
Other than the quantitative comparison, we also visualize point cloud completion results and surface reconstruction results in Figure~\ref{fig:KITTI}, where the mesh is created by Poisson Surface Reconstruction~\cite{kazhdan2013screened}. We can observe that other methods produce more artifacts in the reconstructed surfaces due to the generation of unevenly distributed point set, while our method produce smoother surface and clearer structures in the completion results.

\subsection{Ablation Study and Parameters} 
We demonstrate the effectiveness of the proposed networks through an ablation study, and consider the choice of parameters. We conducted all  experiments on the ShapeNet dataset and all experimental settings including the network architectures were the same as described in~\ref{sec:implementation_details}, except for the analysis.

\begin{figure*}[t!]
\centering
\begin{tabular}{cc}
\begin{tabular}{ccccccc}
\begin{minipage}{.01\textwidth}\centering\raisebox{-.1\height}{ \rotatebox{90}{Input}}\end{minipage} &
\begin{minipage}{.135\textwidth}\centering\raisebox{-.1\height}{\includegraphics[width=\textwidth]{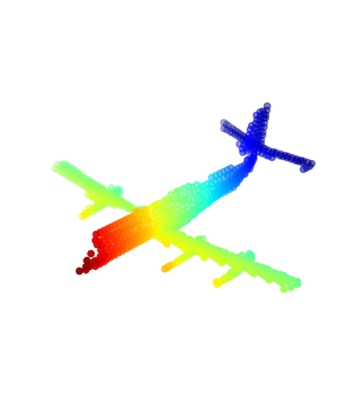}}\end{minipage} &
\begin{minipage}{.135\textwidth}\centering\raisebox{-.1\height}{\includegraphics[width=\textwidth]{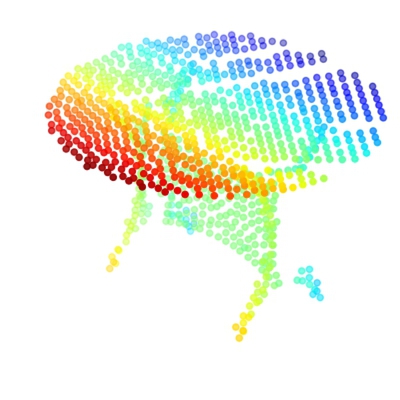}}\end{minipage}&
\begin{minipage}{.135\textwidth}\centering\raisebox{-.1\height}{\includegraphics[width=\textwidth]{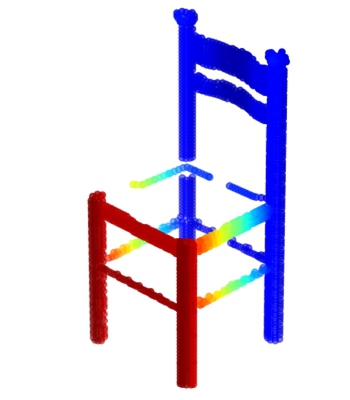}}\end{minipage} &
\begin{minipage}{.135\textwidth}\centering\raisebox{-.1\height}{\includegraphics[width=\textwidth]{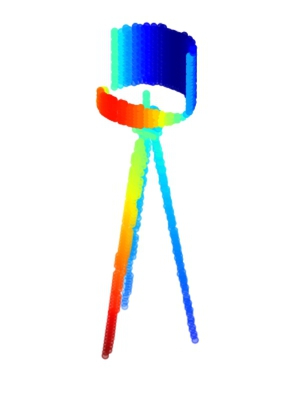}}\end{minipage}&
\begin{minipage}{.135\textwidth}\centering\raisebox{-.1\height}{\includegraphics[width=\textwidth]{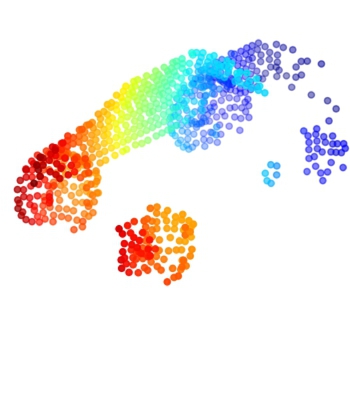}}\end{minipage}&
\begin{minipage}{.135\textwidth}\centering\raisebox{-.1\height}{\includegraphics[width=\textwidth]{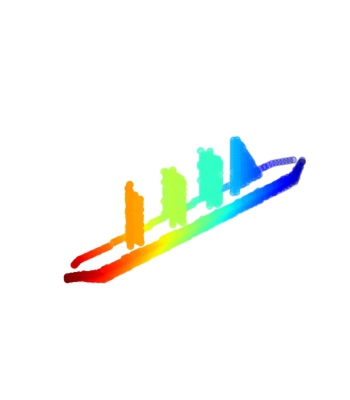}}\end{minipage}
\\
\begin{minipage}{.01\textwidth}\centering\raisebox{-.1\height}{ \rotatebox{90}{GFV}}\end{minipage} &
\begin{minipage}{.135\textwidth}\centering\raisebox{-.1\height}{\includegraphics[width=\textwidth]{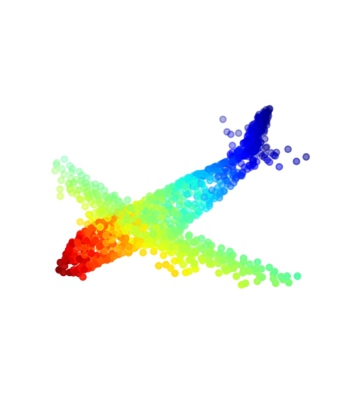}}\end{minipage} &
\begin{minipage}{.135\textwidth}\centering\raisebox{-.1\height}{\includegraphics[width=\textwidth]{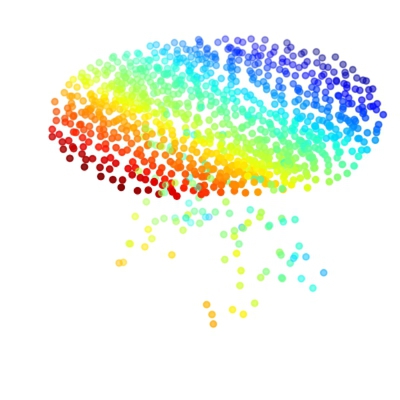}}\end{minipage}&
\begin{minipage}{.135\textwidth}\centering\raisebox{-.1\height}{\includegraphics[width=\textwidth]{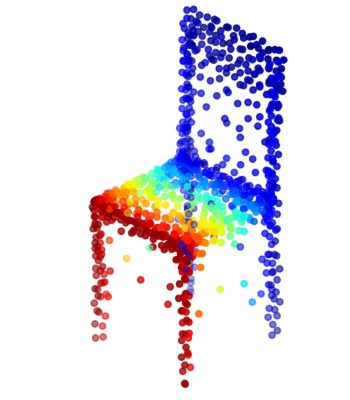}}\end{minipage} &
\begin{minipage}{.135\textwidth}\centering\raisebox{-.1\height}{\includegraphics[width=\textwidth]{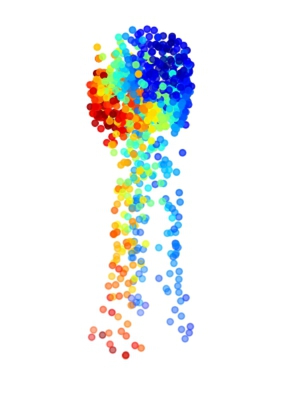}}\end{minipage}&
\begin{minipage}{.135\textwidth}\centering\raisebox{-.1\height}{\includegraphics[width=\textwidth]{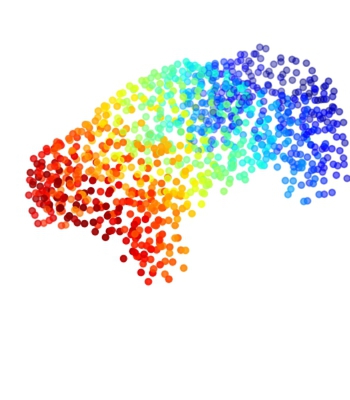}}\end{minipage}&
\begin{minipage}{.135\textwidth}\centering\raisebox{-.1\height}{\includegraphics[width=\textwidth]{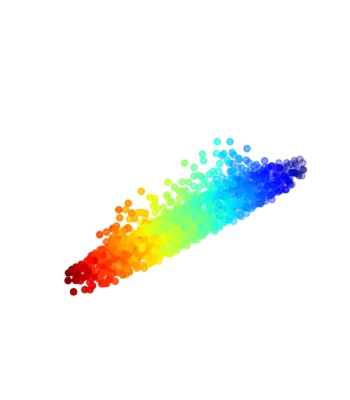}}\end{minipage}
\\
\begin{minipage}{.01\textwidth}\centering\raisebox{-.1\height}{ \rotatebox{90}{SFM}}\end{minipage} &
\begin{minipage}{.135\textwidth}\centering\raisebox{-.1\height}{\includegraphics[width=\textwidth]{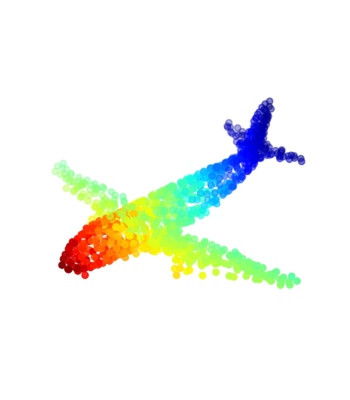}}\end{minipage} &
\begin{minipage}{.135\textwidth}\centering\raisebox{-.1\height}{\includegraphics[width=\textwidth]{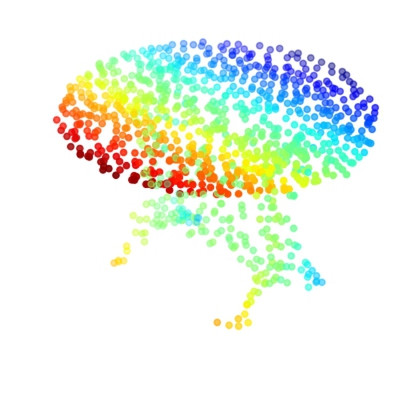}}\end{minipage}&
\begin{minipage}{.135\textwidth}\centering\raisebox{-.1\height}{\includegraphics[width=\textwidth]{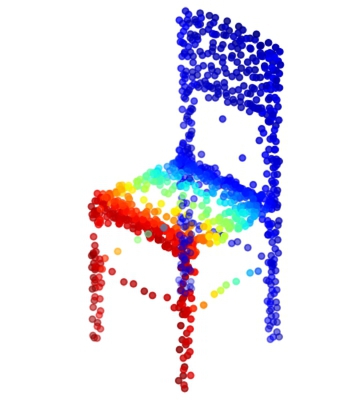}}\end{minipage} &
\begin{minipage}{.135\textwidth}\centering\raisebox{-.1\height}{\includegraphics[width=\textwidth]{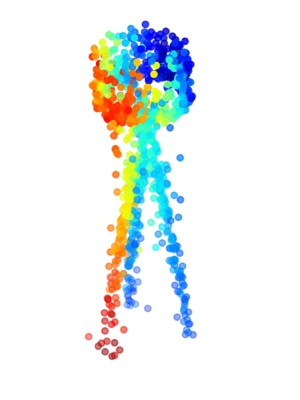}}\end{minipage}&
\begin{minipage}{.135\textwidth}\centering\raisebox{-.1\height}{\includegraphics[width=\textwidth]{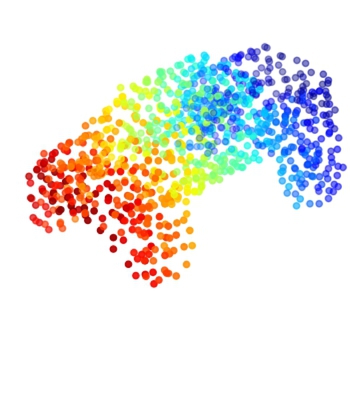}}\end{minipage}&
\begin{minipage}{.135\textwidth}\centering\raisebox{-.1\height}{\includegraphics[width=\textwidth]{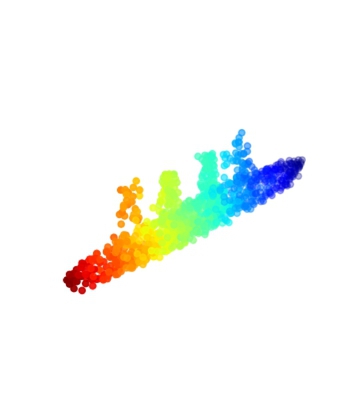}}\end{minipage}
\\
\begin{minipage}{.01\textwidth}\centering\raisebox{-.1\height}{ \rotatebox{90}{GT}}\end{minipage} &
\begin{minipage}{.135\textwidth}\centering\raisebox{-.1\height}{\includegraphics[width=\textwidth]{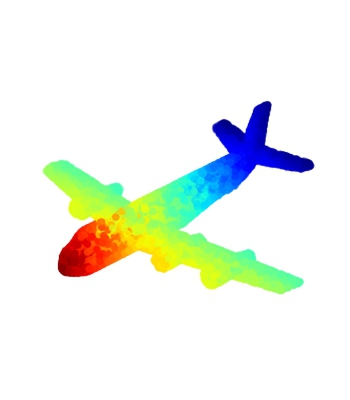}}\end{minipage} &
\begin{minipage}{.135\textwidth}\centering\raisebox{-.1\height}{\includegraphics[width=\textwidth]{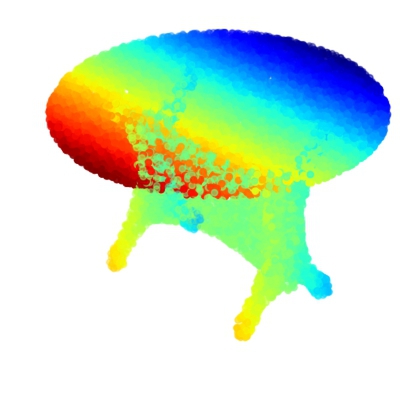}}\end{minipage}&
\begin{minipage}{.135\textwidth}\centering\raisebox{-.1\height}{\includegraphics[width=\textwidth]{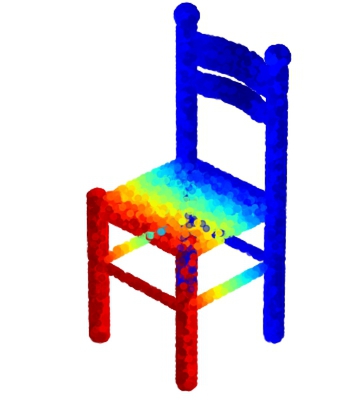}}\end{minipage} &
\begin{minipage}{.135\textwidth}\centering\raisebox{-.1\height}{\includegraphics[width=\textwidth]{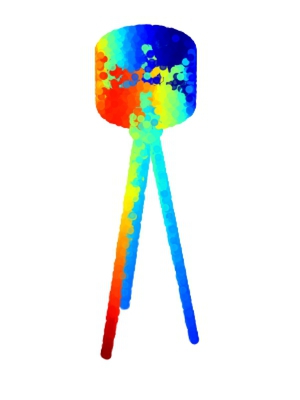}}\end{minipage}&
\begin{minipage}{.135\textwidth}\centering\raisebox{-.1\height}{\includegraphics[width=\textwidth]{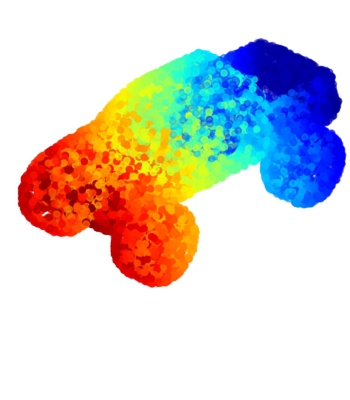}}\end{minipage}&
\begin{minipage}{.135\textwidth}\centering\raisebox{-.1\height}{\includegraphics[width=\textwidth]{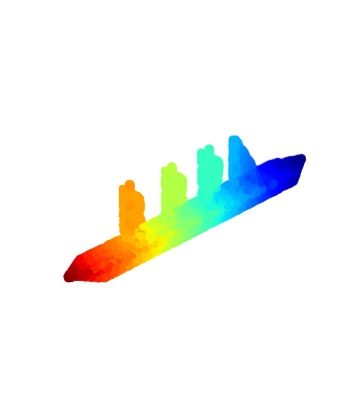}}\end{minipage}
\end{tabular}
\end{tabular}
\caption{Visual comparison of coarse point clouds generated by $\sfm$ and GFV.}
\label{fig:coarse_comparison}
\end{figure*}

\subsubsection{Feature structuring network}
We compared the performance of point cloud completion to quantitatively validate the effect of the FSNet; the structured feature map $\sfm$ was generated with various numbers of channels $h$. Additionally, we replaces  FSNet by a max-pooling operation to produce a global feature vector (GFV) as the global representation instead of $\sfm$, and the coarse point cloud was generated by the module in SnowFlakeNet~\cite{xiang2021snowflakenet}. As  Table~\ref{tab:ablation_FSNet} shows, the completion quality of the $\sfm$-based method for both coarse and dense point clouds is better than for the GFV-based method. A visual comparison of the coarse point clouds generated by different methods is provided in Figure~\ref{fig:coarse_comparison}, and we see that our method provides complete shapes with clearer structures. We also note that CD decreases to the lowest value as $h$ increases to 32, which demonstrates the effectiveness of FSNet. However, CD increases when $h$ rises to 64, which indicates that although using $\sfm$ with more channels can represent richer latent patterns, it may lead to information redundancy. 

\begin{table}[t!]
\centering
\begin{tabular}{l|r|r}
\toprule
\multirow{2}{*}{Method} & \multicolumn{2}{c}{CD} \\ \cline{2-3} 
 & Coarse & Dense \\ \midrule 
GFV           & 9.77 	      & 2.64    \\ \midrule
Ours ($h=8$)  & 7.68	      & 2.40     \\
Ours ($h=16$) & 7.62          & 2.35    \\
Ours ($h=32$) & \textbf{7.26} & \textbf{2.27}  \\  
Ours ($h=64$) & 7.64          & 2.30  \\
\bottomrule
\end{tabular}
\caption{Shape completion (CD loss$\times10^4$) on the ShapeNet dataset for a GFV-based method and ${SFM}$-based method, with various numbers of channels.}
\label{tab:ablation_FSNet}
\end{table}

\begin{figure}
\centering
\renewcommand{\arraystretch}{1.15}
\setlength\tabcolsep{0.5pt}
\begin{tabular}{cc}
\begin{tabular}{ccccc}
 & Airplne & Chair & Lamp & Table \\
\begin{minipage}{.02\textwidth}\centering\raisebox{-.5\height}{\small
\rotatebox{90}{SFM}}\end{minipage} &
\begin{minipage}{.11\textwidth}\centering\raisebox{-.5\height}{\includegraphics[width=\textwidth]{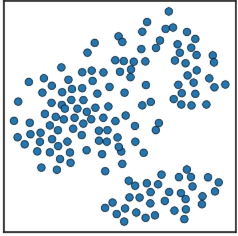}}\end{minipage}&
\begin{minipage}{.11\textwidth}\centering\raisebox{-.5\height}{\includegraphics[width=\textwidth]{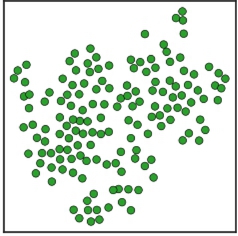}}\end{minipage} &
\begin{minipage}{.11\textwidth}\centering\raisebox{-.5\height}{\includegraphics[width=\textwidth]{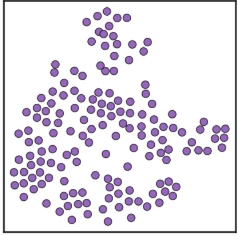}}\end{minipage} &
\begin{minipage}{.11\textwidth}\centering\raisebox{-.5\height}{\includegraphics[width=\textwidth]{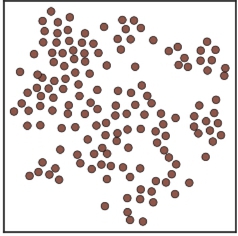}}\end{minipage} \\
\begin{minipage}{.02\textwidth}\centering\raisebox{-.5\height}{\small
\rotatebox{90}{GFV}}\end{minipage} &
\begin{minipage}{.11\textwidth}\centering\raisebox{-.5\height}{\includegraphics[width=\textwidth]{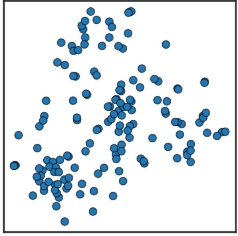}}\end{minipage}& 
\begin{minipage}{.11\textwidth}\centering\raisebox{-.5\height}{\includegraphics[width=\textwidth]{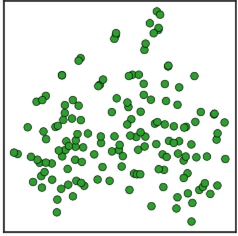}}\end{minipage}&
\begin{minipage}{.11\textwidth}\centering\raisebox{-.5\height}{\includegraphics[width=\textwidth]{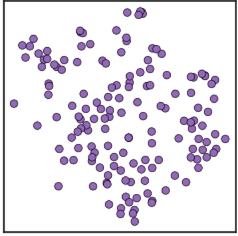}}\end{minipage}&
\begin{minipage}{.11\textwidth}\centering\raisebox{-.5\height}{\includegraphics[width=\textwidth]{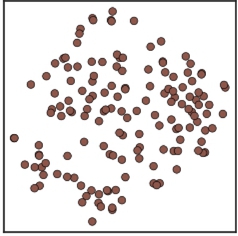}}\end{minipage}
\end{tabular}
\end{tabular}
\caption{Compared to GFV used in~\cite{xiang2021snowflakenet}, the feature embedding of the learned $\mathit{SFM}$ is more uniformly distributed. }
\label{fig:feature_distribution}
\end{figure}

We also provide a visualization of the learned $\sfm$ projection for various testing samples from the ShapeNet dataset in Figure~\ref{fig:feature_distribution}. Compared to a recent GFV based method~\cite{xiang2021snowflakenet}, the distribution of $\sfm$ projections is more uniform at instance level across different categories, demonstrating that $\sfm$ has learned sufficient local information to distinguish different shapes. Note that we use the same encoder as~\cite{xiang2021snowflakenet} to conduct the comparison.

\begin{table*}[t!]
\centering
\renewcommand{\arraystretch}{1.15}
\begin{tabular}{l|rrrrrrrr|r}
\toprule
\multicolumn{1}{c|}{Method} & Plane      & Cabinet       & Car           & Chair         & Lamp          & Sofa          & Table         & Vessel    & Avg.       \\ \midrule \midrule
MB~\cite{yu2018pu}  & 0.99	         & 3.80	         & 2.30	         & 3.64	         & 3.12	         & 4.26	         & 2.77	         & 2.61         & 2.94          \\
DP~\cite{li2019pu}  & 1.00	         & 3.88	         & 2.30	         & 3.46	         & 3.07	         & 4.24	         & 2.75	         & 2.60         & 2.92          \\
NS~\cite{qian2021pu} & 0.96	         & 3.79	         & 2.28	         & 3.23	         & 2.95	         & 4.17	         & 2.68	         & 2.51         & 2.82          \\ \midrule
Ours ($T=1$) & 0.80          & 3.64          & 2.21          & 3.05          & 2.23          & 3.87          & 2.31          & 1.88 & 2.50          \\
Ours ($T=3$) & \textbf{0.79} & 3.45          & 2.24          & 2.86          & 1.99          & 3.98          & 2.27          & 2.03          & 2.45          \\
Ours ($T=5$) & 0.82          & 3.52          & 2.18          & 2.97          & 1.85          & 3.65 & 2.12          & 1.92          & 2.38          \\
Ours ($T=7$) & 0.84          & \textbf{3.47}          & 2.18          & 2.63 & 1.84 & 3.73          & 2.04 & 1.90          & 2.33 \\
Ours ($T=9$) & 0.80          & 3.53 & \textbf{2.13} & \textbf{2.48}          & \textbf{1.79}          & \textbf{3.64}          & \textbf{1.98}          & \textbf{1.81}          & \textbf{2.27}          \\ 
\bottomrule
\end{tabular}
\caption{Shape completion results (CD loss$\times 10^4$) on the ShapeNet dataset with different feature expansion networks.}
\label{tab:ablation_shapenet}
\end{table*}

\begin{figure*}[t!]
\centering
\begin{tabular}{cccccc}
Input & MB & DP & NS & Ours & GT \\
\includegraphics[width=0.14\textwidth]{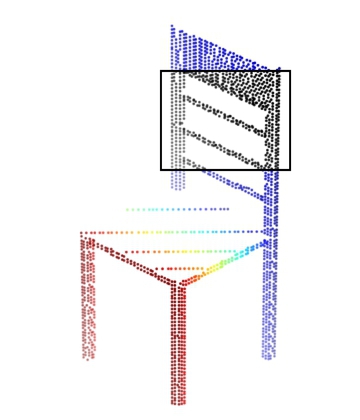} & 
\includegraphics[width=0.14\textwidth]{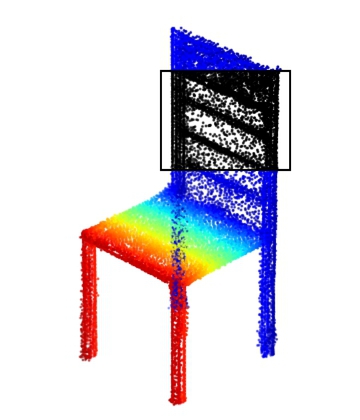} & 
\includegraphics[width=0.14\textwidth]{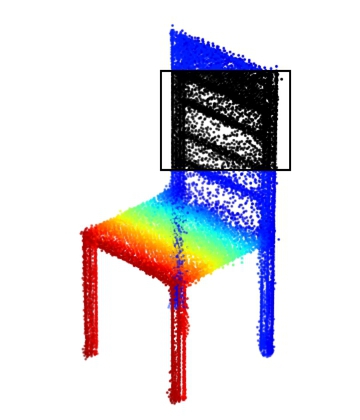} & 
\includegraphics[width=0.14\textwidth]{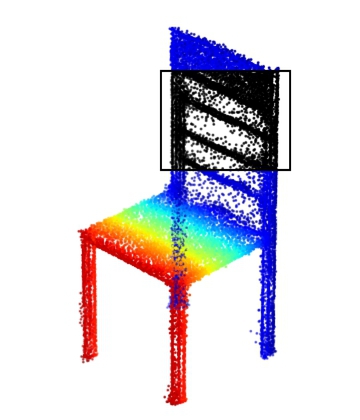} & 
\includegraphics[width=0.14\textwidth]{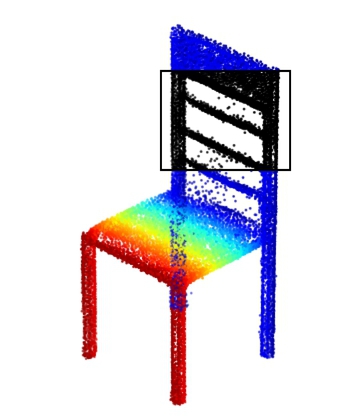} & 
\includegraphics[width=0.14\textwidth]{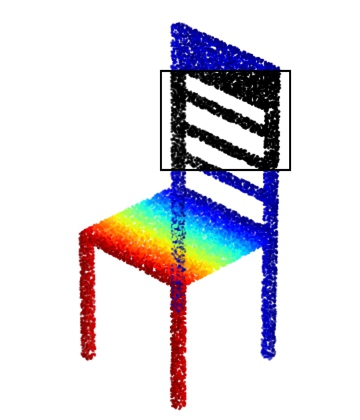}
\\
\includegraphics[width=0.14\textwidth]{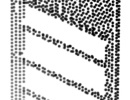} & 
\includegraphics[width=0.14\textwidth]{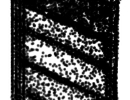} & 
\includegraphics[width=0.14\textwidth]{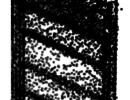} & 
\includegraphics[width=0.14\textwidth]{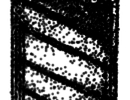} & 
\includegraphics[width=0.14\textwidth]{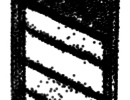} & 
\includegraphics[width=0.14\textwidth]{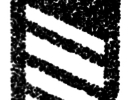}
\\  
\includegraphics[width=0.14\textwidth]{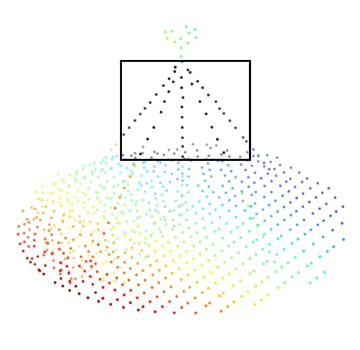} & 
\includegraphics[width=0.14\textwidth]{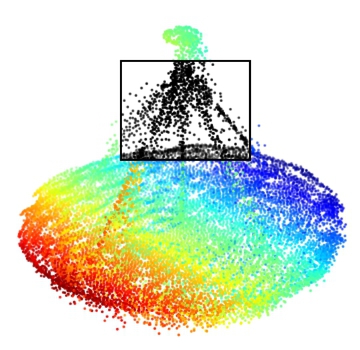} & 
\includegraphics[width=0.14\textwidth]{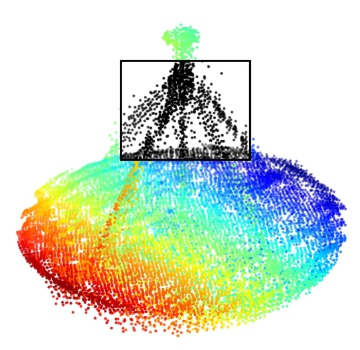} & 
\includegraphics[width=0.14\textwidth]{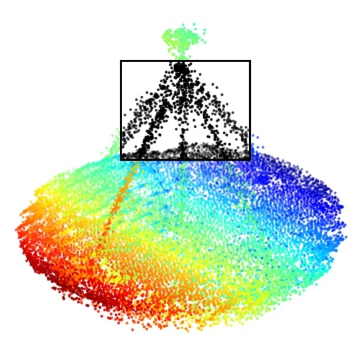} & 
\includegraphics[width=0.14\textwidth]{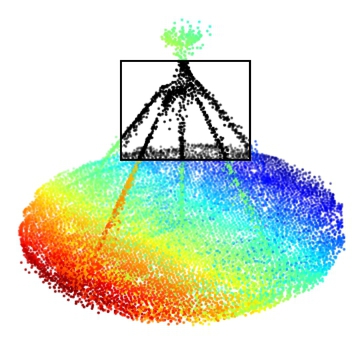} & 
\includegraphics[width=0.14\textwidth]{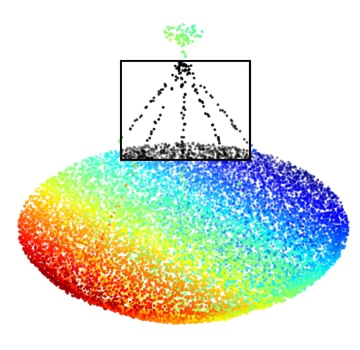}
\\
\includegraphics[width=0.14\textwidth]{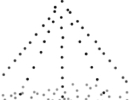} & 
\includegraphics[width=0.14\textwidth]{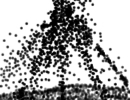} & 
\includegraphics[width=0.14\textwidth]{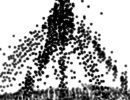} & 
\includegraphics[width=0.14\textwidth]{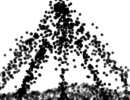} & 
\includegraphics[width=0.14\textwidth]{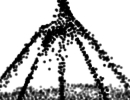} & 
\includegraphics[width=0.14\textwidth]{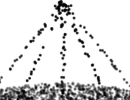}
\end{tabular}
\caption{Example shape completion results using the ShapeNet dataset using different feature expansion networks.}
\label{fig:ablation_shapenet}
\end{figure*}

\subsubsection{Iterative feedback network}
Feature expansion is an important component of recent point cloud upsampling methods. However, existing feature expansion methods, including multi-branch MLPs~(MB)~\cite{yu2018pu}, duplication-based expansion~(DP)~\cite{li2019pu} and NodeShuffle~(NS)~\cite{qian2021pu} generally perform expansion in one or two steps, which limits their abilities in the point cloud completion task which has incomplete and non-uniform inputs. To demonstrate the effectiveness of IFNet, we conducted a comparison in which IFNet in our pipeline is either replaced by other feature expansion methods or has various numbers $T$ of feedback blocks. Quantitative results are shown in Table~\ref{tab:ablation_shapenet}, which demonstrates the advantage of IFNet over other methods in the point cloud completion task; the completion quality  improves as the number of feedback blocks increases. Qualitative completion results and close-ups are illustrated in Figure~\ref{fig:ablation_shapenet}, showing that the dense point clouds generated by our method have fewer outliers and less noise, demonstrating that the proposed IFNet facilitates consolidation and uniformity of output point clouds.

\begin{figure*}[t!]
\centering
\begin{tabular}{cccccc}
$t=1$ & $t=2$ & $t=3$ & $t=4$ & $t=5$ & GT \\
\includegraphics[width=0.14\textwidth]{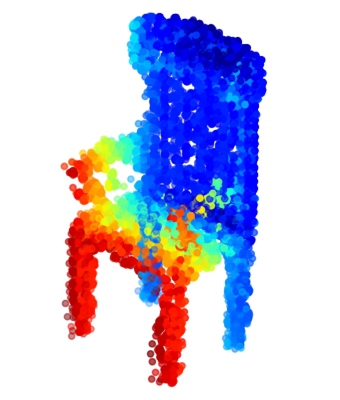} & 
\includegraphics[width=0.14\textwidth]{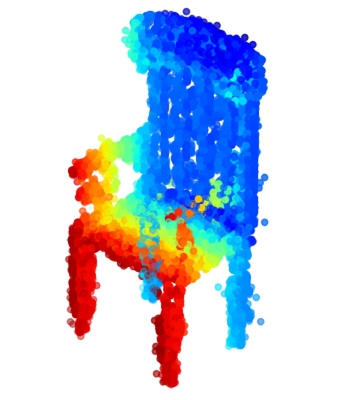} & 
\includegraphics[width=0.14\textwidth]{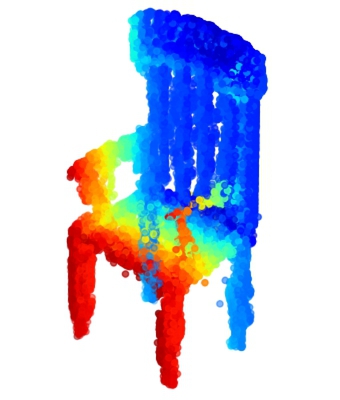} & 
\includegraphics[width=0.14\textwidth]{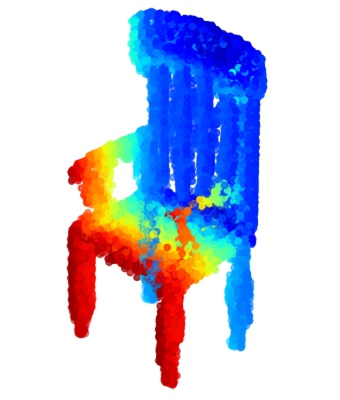} & 
\includegraphics[width=0.14\textwidth]{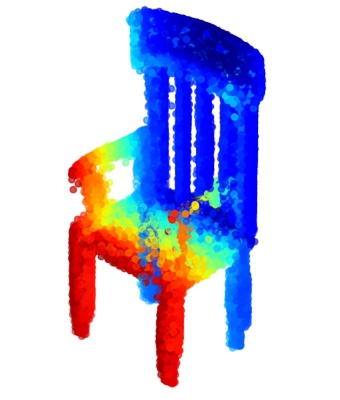} & 
\includegraphics[width=0.14\textwidth]{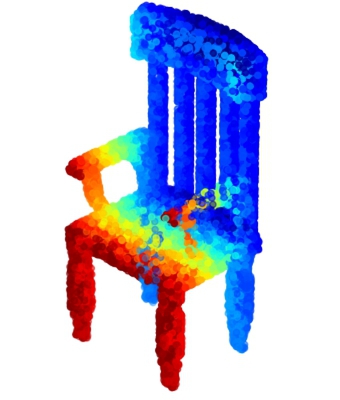}
\\
\includegraphics[width=0.14\textwidth]{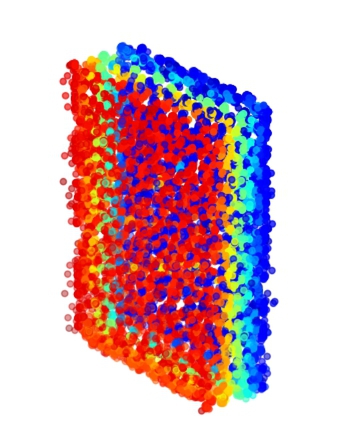} & 
\includegraphics[width=0.14\textwidth]{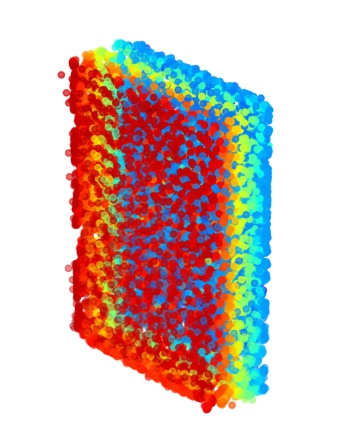} & 
\includegraphics[width=0.14\textwidth]{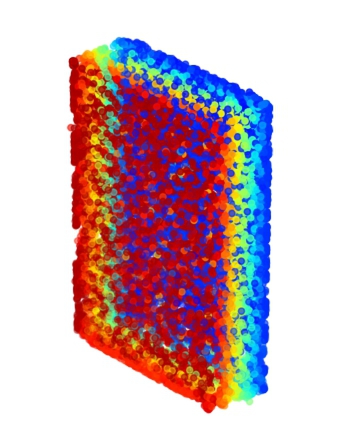} & 
\includegraphics[width=0.14\textwidth]{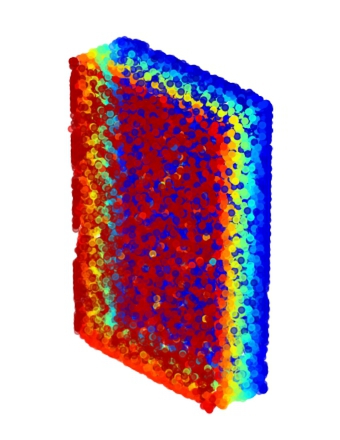} & 
\includegraphics[width=0.14\textwidth]{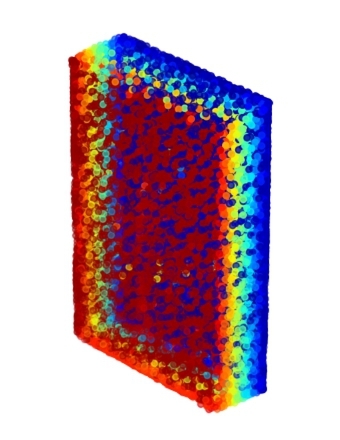} & 
\includegraphics[width=0.14\textwidth]{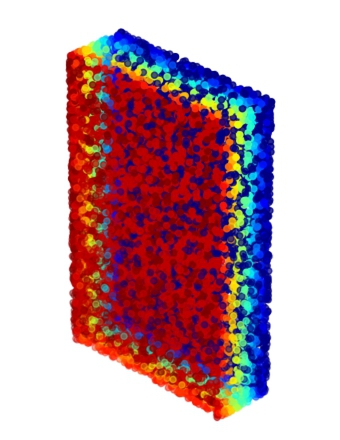}
\end{tabular}
\caption{Example shape completion results using the ShapeNet dataset regressed from the intermediate features in IFNet.}
\label{fig:IFNet_Itermediate}
\end{figure*}

For more comprehensive understanding of the progressive refinement process in IFNet, we visualize various intermediate completion results in Figure~\ref{fig:IFNet_Itermediate}, where the offsets between the coordinates of points in ${P^s}$ and the dense point cloud in step $t$ are regressed from the corresponding dense features ${F^d_t}$. It shows that as the the dense features are progressively corrected via the feedback mechanism, local details and surface uniformity of the dense point cloud are gradually improved. Note that only the dense features from the last feedback block are fed to the offset regression module during both  training and test stages, and there is no need to supervise the intermediate results as in previous methods~\cite{xiang2021snowflakenet, huang2020pf}.

\section{Conclusions}
In this study, we revisit the problem of point cloud completion and propose two novel networks which can be integrated into a coarse-to-fine pipeline. By replacing the max pooling operation with FSNet, we are able to efficiently aggregate both global and local information from  partial observations. Moreover, we introduce IFNet into the upsampling stage, which can work in a self-correcting manner, helping to progressively refine local details of the output shape. Experiments on multiple datasets indicate that our method achieves state-of-the-art performance. 

However, our method still has certain limitations. FSNet incurs an extra computational cost compared to max-pooing, and it may lead to information redundancy as noted in the parameter study. Furthermore, IFNet takes multiple steps to expand the sparse features, which is also relatively time-consuming. It is worth exploring ways to address these limitations and the possibility of applying our method to other tasks like point cloud consolidation and upsampling. 
Moreover, since we use multi-head attention in FSNet and self-attention in IFNet, it would be interesting to explore other more sophisticated attention mechanisms~\cite{guo2021attention}, \eg transformer-based modules that have  proved helpful in 3D point cloud processing~\cite{guo2021pct, zhao2021point}, to see if the results can be further improved.

{\small
\bibliographystyle{cvm}
\bibliography{cvmbib}
}

\end{document}